\documentclass[twocolumn,journal]{IEEEtran}
\usepackage[latin9]{inputenc}
\PassOptionsToPackage{ruled}{algorithm2e}
\usepackage{color}
\usepackage{algorithm2e}
\usepackage{amsmath}
\usepackage{amssymb}
\usepackage{graphicx}
\PassOptionsToPackage{normalem}{ulem}
\usepackage{ulem}
\usepackage{subscript}
\usepackage[unicode=true,
 bookmarks=true,bookmarksnumbered=true,bookmarksopen=true,bookmarksopenlevel=1,
 breaklinks=false,pdfborder={0 0 0},pdfborderstyle={},backref=false,colorlinks=false]
 {hyperref}
\hypersetup{pdftitle={Your Title},
 pdfauthor={Your Name},
 pdfpagelayout=OneColumn, pdfnewwindow=true, pdfstartview=XYZ, plainpages=false}

\makeatletter
\usepackage{inputenc}

\@ifundefined{showcaptionsetup}{}{%
 \PassOptionsToPackage{caption=false}{subfig}}
\usepackage{subfig}
\makeatother

\begin{document}

\title{Multiple Instance Learning with the Optimal Sub-Pattern Assignment
Metric}

\author{Quang N. Tran, Ba-Ngu Vo, Dinh Phung, Ba-Tuong Vo, and Thuong Nguyen}
\maketitle
\begin{abstract}
Multiple instance data are sets or multi-sets of unordered elements.
Using metrics or distances for sets, we propose an approach to several
multiple instance learning tasks, such as clustering (unsupervised
learning), classification (supervised learning), and novelty detection
(semi-supervised learning). In particular, we introduce the Optimal
Sub-Pattern Assignment metric to multiple instance learning so as
to provide versatile design choices. Numerical experiments on both
simulated and real data are presented to illustrate the versatility
of the proposed solution.  
\end{abstract}


\IEEEpeerreviewmaketitle{}

\section{Introduction\label{sec:Introduction} }

Multiple instance (MI) data, more commonly known as `bags' \cite{dietterich1997solving_multiple_instance},
\cite{minhas2012multiple}, \cite{amores2013multiple_intance_review,foulds2010multi_instance_review},
are mathematical objects called point patterns. A point pattern (PP)
is a set or multi-set of unordered points (or elements) \cite{moller2003point_processes},
in which each point represents the state or features of the object
of study. Note that a set does not contain repeated points while a
multi-set can. PPs appear in a variety of applications. In natural
language processing and information retrieval, the `bag-of-words'
representation treats each document as a collection or set of words
\cite{joachims1996probabilistic,mccallum1998comparison_NBtextClassifi}.
In image and scene categorization, the `bag-of-visual-words' representation\textemdash the
analogue of the `bag-of-words' in text analysis\textemdash treats
each image as a set of its key patches \cite{csurka2004visual,fei2005bayesian}.
In applications involving three-dimensional (3D) images such as computer
tomography scan, and magnetic resonance imaging, point cloud data
are actually sets of points in some coordinate system \cite{rusu2011point_cloud_lib,sitek2006tomographic_point_cloud,woo2002segmentation_point_cloud}.
In data analysis for the retail industry as well as web management
systems, transaction records such as market-basket data \cite{guha1999rock,yang2002clope,yun2001clustering_basket_data}
and web log data \cite{cadez2000EMclustering_VariableLengthData}
are sets of transaction items.  

While PP data are abundant, fundamental MI learning tasks such as
clustering (unsupervised learning), classification (supervised learning),
and novelty detection\footnote{Novelty detection is not a special case of classification because
anomalous or novel training data is not available \cite{hodge2004survey}.} (semi-supervised learning), have received limited attention \cite{amores2013multiple_intance_review,foulds2010multi_instance_review}.
Indeed, to the best of our knowledge, there are no MI learning solutions
based on PP models, nor any MI novelty detection solutions in the
literature. 

In MI clustering, two algorithms have been developed for PP data:
Bag-level Multi-instance Clustering (BAMIC) \cite{zhang2009MIClustering};
and Maximum Margin Multiple Instance Clustering (M$^{3}$IC) \cite{zhang2009m3icClustering}.
BAMIC adapts the $k$-medoids algorithm with the Hausdorff distance
as a measure of dissimilarity between PPs \cite{zhang2009MIClustering}.
M$^{3}$IC, on the other hand, poses the PP clustering problem as
a non-convex optimization problem which is then relaxed and solved
via a combination of the Constrained Concave-Convex Procedure and
Cutting Plane methods \cite{zhang2009m3icClustering}. 

In MI classification, there are three paradigms: Instance-Space; Embedded-Space;
and Bag-Space \cite{amores2013multiple_intance_review,foulds2010multi_instance_review}.
These paradigms differ in the way they exploit data at the local level
(individual points within each bag) or at the global level (the bags
themselves as observations). Instance-Space is the only paradigm exploiting
data at the local level which neglect the relationship between points
in the bag. At the global level, the Embedded-Space paradigm maps
all PPs to vectors of fixed dimension, which are then processed by
standard classifiers for vectors. On the other hand, the Bag-Space
paradigm addresses the problem at the most fundamental level by operating
directly on the PPs. The philosophy of the Bag-Space paradigm is to
preserve the information content of the data, which could otherwise
be compromised through the data transformation process (as in the
Embedded-Space approach). Existing methods in the Bag-Space paradigm
uses the Hausdorff \cite{huttenlocher1993tracking_Hausdorff}, Chamfer
\cite{gavrila1999Chamfer_distance}, and Earth Mover's \cite{zhang2007EMD_kernel,rubner1998EarthMoversDistance}
distances. 

In this paper, we propose the use of the Optimal Sub-Pattern Assignment
(OSPA) distance \cite{schuhmacher2008_OSPA} in MI clustering, classification
and novelty detection. The choice of set distance in MI learning can
markedly influence the performance, and the OSPA distance provides
more flexible design choices for different types of applications.
Our specific contributions are:
\begin{itemize}
\item In MI clustering, we combine the Affinity Propagation (AP) clustering
algorithm \cite{frey2007_APclustering} with set distances as dissimilarity
measures\footnote{Preliminary results have been presented in the conference paper \cite{tran2016clustering_PP}.
This paper presents a more comprehensive study.}. We also examine the clustering performance amongst the OSPA, Hausdorff,
and Wasserstein distances. Compared to existing $k$-medoids based
techniques \cite{zhang2009MIClustering}, AP can find clusters faster
with much lower error, and does not require the number of clusters
to be specified \cite{frey2007_APclustering}. In addition, the OSPA
distance is more versatile than the Hausdorff distance used in \cite{zhang2009MIClustering}.
\item In MI classification, we use the OSPA distance in the $k$-nearest
neighbour ($k$-NN) algorithm \cite{cover1967nearest_neighbor_classifier,keller1985fuzzy_kNN},
and examine the performance against the Hausdorff-based technique
\cite{huttenlocher1993tracking_Hausdorff} and the Wasserstein-based
technique (the Earth Mover's distance adapted for PPs \cite{hoffman2004multitarget_distance}).
Being a Bag-Space approach, this technique exploits data at the global
level, and avoids potential information loss from the embedding. Moreover,
the advantage over existing Bag-Space approaches lies in the versatility
of the OSPA distance over the Hausdorff \cite{huttenlocher1993tracking_Hausdorff},
Chamfer \cite{gavrila1999Chamfer_distance} and Earth mover's \cite{zhang2007EMD_kernel}
distances. 
\item In MI novelty detection, we propose a solution based on the set distance
between the candidate PP and its nearest neighbour in the normal training
set. We also examine the detection performance amongst the OSPA, Hausdorff,
and Wasserstein distances. This very first MI novelty detection method
is simple, effective and versatile across various applications. 
\end{itemize}
The rest of this paper is organized as follows. Section \ref{sec:Set-distances}
presents the Hausdorff, Wasserstein and OSPA distances along with
their properties in the context of MI. Based on these distances, sections
\ref{sec:Clustering}, \ref{sec:Classification}, and \ref{sec:Novelty-detection}
present the distance-based MI learning algorithms and numerical experiments
for clustering, classification, and novelty detection, respectively.
Section \ref{sec:Conclusions} concludes the paper. 

\section{Set distances\label{sec:Set-distances}}

Machine learning tasks such as clustering, classification and novelty
detection are mainly concerned with the grouping/separating of data
based on their similarities/dissimilarities. A distance is a fundamental
measure of dissimilarity between two objects. Hence, the notion of
distance or metric is important to learning approaches without models
\cite{jain2010clustering50yearsKmeans,amores2013multiple_intance_review,pimentel2014review_novel_detect}.
In MI learning, several set distances have been introduced for PP
data\footnote{A multi-set can be equivalently expressed as a set by augmenting the
multiplicity of each element, i.e., a multi-set with elements $x_{1}$
repeated $N_{1}$ times, ...., $x_{m}$ repeated $N_{m}$ times, can
be represented as the set $\{(x_{1},N_{1}),...,(x_{m},N_{m})\}$.}, namely the Hausdorff \cite{huttenlocher1993tracking_Hausdorff},
and Chamfer \cite{gavrila1999Chamfer_distance} distances.

In this section, we present the Hausdorff \cite{huttenlocher1993tracking_Hausdorff},
Wasserstein \cite{hoffman2004multitarget_distance}, and OSPA distances
\cite{schuhmacher2008_OSPA}. In particular we discuss their properties
and the implications in the context of design choices for MI learning
algorithms. The choice of set distance in MI learning directly influences
the performance and hence it is important to select distances that
are compatible with the applications. 

For completeness, we recall the definition of a distance function
or metric on a non-empty space $\mathcal{S}$. A function $d:\mathcal{S\times}\mathcal{S}\rightarrow[0;1)$
is called a metric if it satisfies the following three axioms: 
\begin{enumerate}
\item (Identity) $d(x,y)=0$ if and only if $x=y$ ; 
\item (Symmetry) $d(x,y)=d(y,x)$ for all $x,y\in\mathcal{S}$ ; 
\item (Triangle inequality) $d(x,y)\leq d(x,z)+d(z,y)$ for all $x,y,z\in\mathcal{S}$.
\end{enumerate}
Our interest lies in the distance between two finite subsets $X=\{x_{1},...,x_{m}\}$
and $Y=\{y_{1},...,y_{n}\}$ of a metric space $(\mathcal{W},\underline{d})$,
where $\mathcal{W}$ is closed and bounded observation window, and
$\underline{d}$ denotes \emph{the base distance} between the elements
of $\mathcal{W}$. Note that $\underline{d}$ is usually taken as
the Euclidean distance when $\mathcal{W}$ is a subset of $\mathbb{R}^{n}$.

\subsection{Hausdorff distance\label{subsec:Hausdorff-distance}}

The Hausdorff distance between two non-empty sets $X$ and $Y$ is
defined by\vspace{-1mm}
\begin{equation}
d_{\mathtt{H}}(X,Y)=\max\left\{ \max_{x\in X}\min_{y\in Y}\underline{d}(x,y),\max_{y\in Y}\min_{x\in X}\underline{d}(x,y)\right\} ,\label{eq:Hausdorff_dist}
\end{equation}
Note that the Hausdorff distance is not defined when either $X$ or
$Y$ is empty. 

In addition to being a metric, the Hausdorff distance is easy to compute
and was traditionally used as a measure of dissimilarity between binary
images. It gives a good indication of the dissimilarity in the visual
impressions that a human would typically perceive between two binary
images. Hausdorff distance has been successfully applied in applications
dealing with PP data, such as detecting objects from binary images
\cite{huttenlocher1993comparing,rucklidge1995locating_obj}, or measuring
the dissimilarities between 3-D surfaces\textemdash sets of coordinates
of points \cite{cignoni1998metro}. In MI learning it has been applied
in classification \cite{amores2013multiple_intance_review} and clustering
\cite{zhang2009MIClustering}.
\begin{flushleft}
\vspace{-1mm}
\begin{figure}[tbh]
\noindent \begin{centering}
\begin{minipage}[c]{48mm}%
\begin{center}
\includegraphics[width=1\columnwidth]{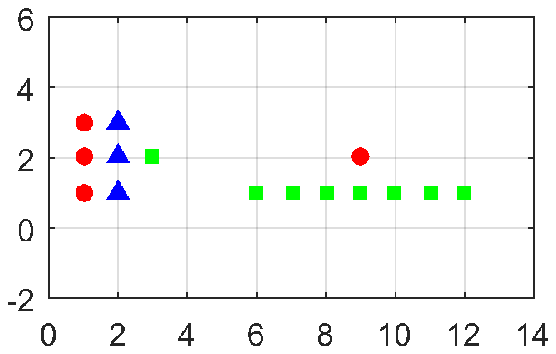}
\par\end{center}%
\end{minipage}\hspace{3mm}%
\begin{minipage}[c]{35mm}%
\begin{center}
\includegraphics[width=1\columnwidth]{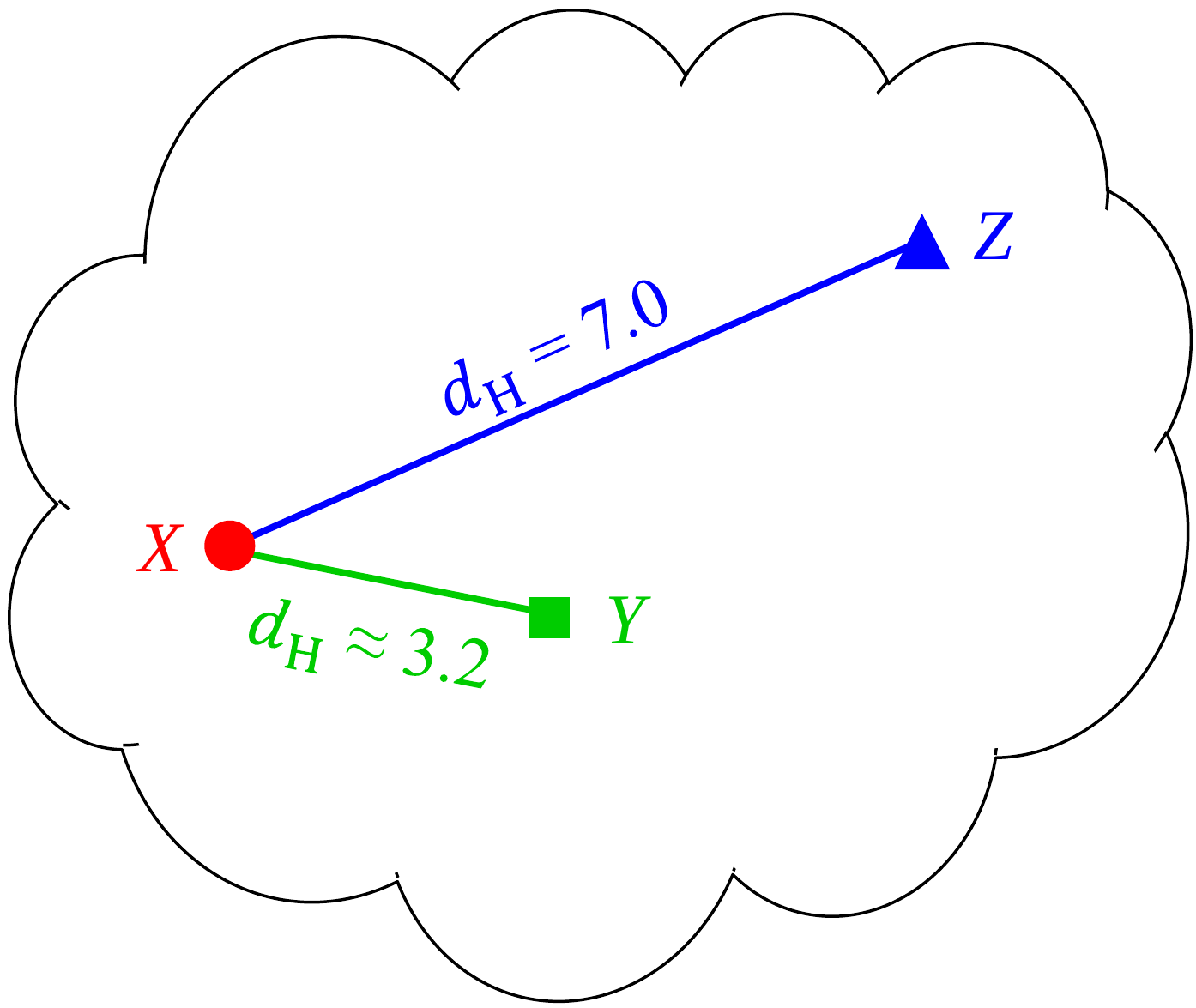}
\par\end{center}%
\end{minipage}
\par\end{centering}
\begin{centering}
\caption{\label{fig:Hausdorff_demo} Cardinality difference and out-liers in
the Hausdorff distance. Left: Sets $X$ (red ${\color{red}\bullet}$),
$Y$ (green {\tiny{}${\normalcolor {\color{green}\blacksquare}}$}),
and $Z$ (blue{\footnotesize{} ${\color{blue}\blacktriangle}$}) in
$\mathbb{R}^{2}$. Right: Abstract impression of the Hausdorff distances
between the finite sets $X$, $Y$, and $Z$.}
\par\end{centering}
\vspace{-2mm}
\end{figure}
\par\end{flushleft}

The Hausdorff distance could produce some undesirable effects for
many MI learning applications since it may group together PPs that
are intuitively dissimilar while separating PPs that are similar.
Specifically:
\begin{itemize}
\item The Hausdorff distance is relatively insensitive to dissimilarities
in cardinality \cite{schuhmacher2008_OSPA}. Consequently, it can
group together PPs with large differences in cardinality (e.g., $X$
and $Y$ in Fig. \ref{fig:Hausdorff_demo}). This can be undesirable
in many applications since the cardinalities of the PPs are important
in MI learning. 
\item The Hausdorff distance penalizes heavily outliers\textemdash elements
in one set which are far from every element of the other set \cite{schuhmacher2008_OSPA}.
Consequently, it tends to separate similar sets that differ only in
a few outliers (e.g., $X$ and $Z$ in Fig. \ref{fig:Hausdorff_demo}).
This is undesirable in applications where the observed PPs of underlying
groups are contaminated by outliers due to spurious noise. Nonetheless,
there are applications where it is desirable to separate PPs with
outliers from those without. Note that there are also generalizations
of the Hausdorff distance that avoid the undesirable outlier penalty
\cite{baddeley1992errors}.
\end{itemize}
The Chamfer ``distance'' \cite{gavrila1999Chamfer_distance} is
a variation of the Hausdorff construction, but does not satisfy the
metric axioms. In terms of measuring dissimilarity, it is very similar
to the Hausdorff distance, and has been used to construct a Support
Vector Machine kernel for MI classification in \cite{amores2013multiple_intance_review}.

\subsection{Wasserstein distance\label{subsec:Wasserstein-distance}}

The Wasserstein distance (also known as Optimal Mass Transfer distance
\cite{schuhmacher2008_OSPA}) of order $p\geq1$ between two sets
$X$ and $Y$ is defined by \cite{hoffman2004multitarget_distance}
\vspace{-1mm}
\begin{equation}
d_{\mathtt{W}}^{(p)}(X,Y)=\min_{C}\left(\sum_{i=1}^{m}\sum_{j=1}^{n}c_{i,j}\underline{d}\left(x_{i},y_{j}\right)^{p}\right)^{\frac{1}{p}},\label{eq:OMAT-dist}
\end{equation}
where $C=\left(c_{i,j}\right)$ is an $m\times n$ transportation
matrix (recall that $m$ and $n$ are the cardinalities of $X$ and
$Y$, respectively), i.e., $c_{i,j}$ are non-negative and satisfies:\begin{subequations}\label{eq:trans_mat_criteria}
\begin{alignat}{1}
\sum_{j=1}^{n}c_{i,j}=\frac{1}{m} & \,\mbox{for }1\leq i\leq m,\\
\sum_{i=1}^{m}c_{i,j}=\frac{1}{n} & \,\mbox{for }1\leq j\leq n.
\end{alignat}
\end{subequations}Note that similar to the Hausdorff distance the
Wasserstein distance is a metric \cite{hoffman2004multitarget_distance}
and is not defined when either $X$ or $Y$ is empty. 

\begin{figure}[tbh]
\begin{centering}
\begin{minipage}[c]{50mm}%
\begin{center}
\includegraphics[width=1\columnwidth]{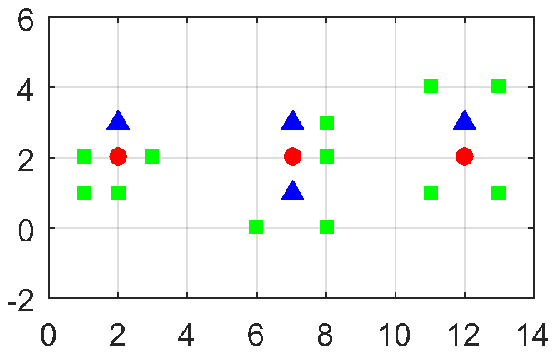}
\par\end{center}%
\end{minipage}\hspace{3mm}%
\begin{minipage}[c]{35mm}%
\begin{center}
\includegraphics[width=1\columnwidth]{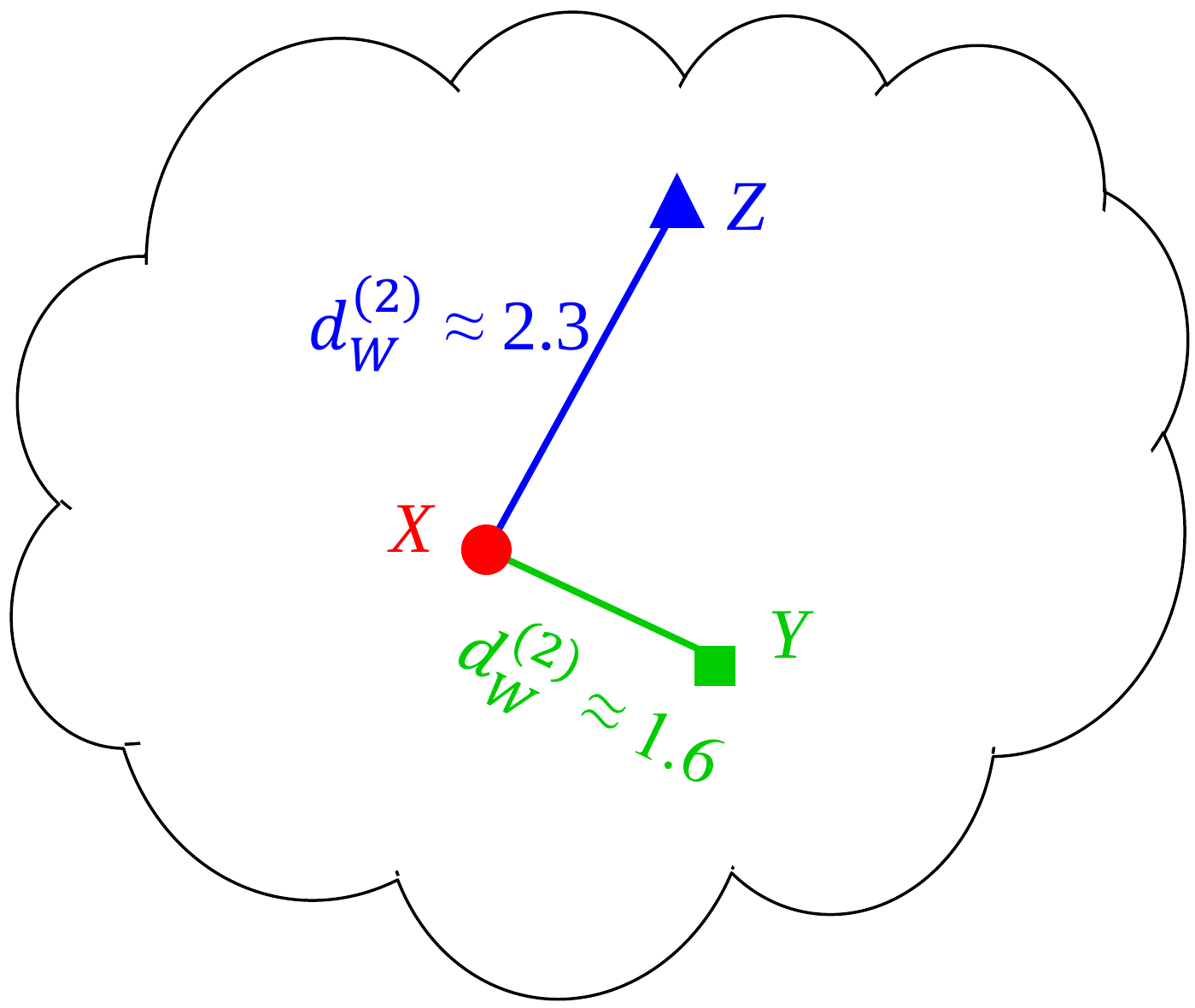}
\par\end{center}%
\end{minipage}
\par\end{centering}
\caption{\label{fig:Wasserstein_demo}Left: Sets $X$ (red ${\color{red}\bullet}$),
$Y$ (green {\tiny{}${\normalcolor {\color{green}\blacksquare}}$}),
and $Z$ (blue{\footnotesize{} ${\color{blue}\blacktriangle}$}) in
$\mathbb{R}^{2}$. Right: Abstract impression of the Wasserstein distances
between the finite sets $X$, $Y$, and $Z$.}
\end{figure}

\begin{figure}[tbh]
\begin{centering}
\vspace{-1mm}
\subfloat[To compute \textbf{$d_{\mathtt{W}}^{(p)}(X,Y)$}: move $Y$'s earth
piles (green) to form $X$'s earth piles (red).]{\begin{centering}
\includegraphics[width=60mm]{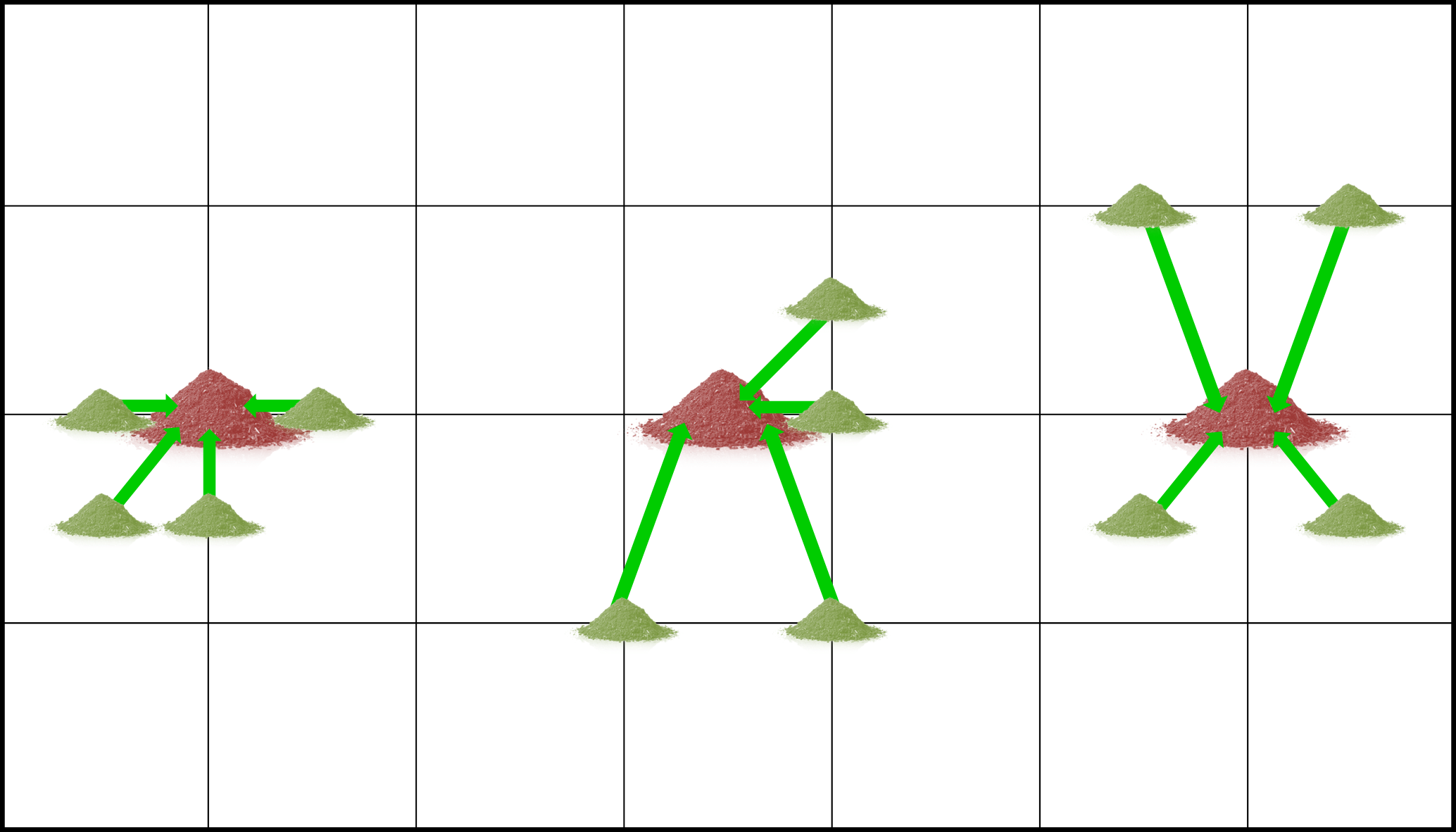}
\par\end{centering}
}
\par\end{centering}
\begin{centering}
\subfloat[\label{fig:EarthMoversDistance-b}To compute \textbf{$d_{\mathtt{W}}^{(p)}(X,Z)$}:
move $Z$'s earth piles (blue) to form $X$'s earth piles (red).]{\begin{centering}
\includegraphics[width=60mm]{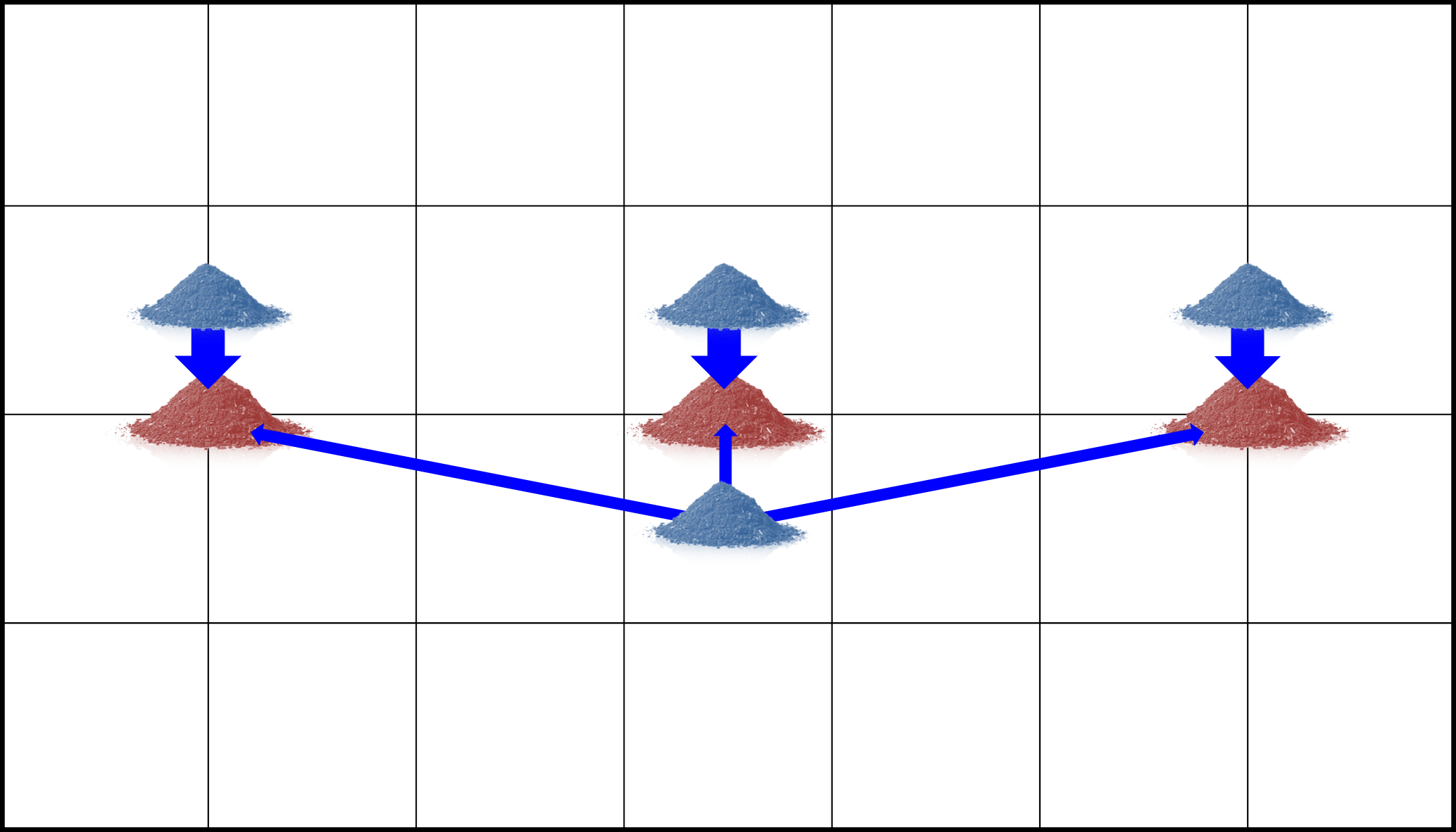}
\par\end{centering}
}\vspace{-1mm}
\par\end{centering}
\caption{\label{fig:EarthMoversDistance} Earth mover's interpretation of the
Wasserstein distance. Sets $X$, $Y$, and $Z$ in Fig. \ref{fig:Wasserstein_demo}
considered as collections of earth piles. The blue/green arrows represent
the amounts and directions of transportation of the blue/green earth
piles.}
\end{figure}

The Wasserstein distance can be considered as the Earth Mover's distance
\cite{rubner1998EarthMoversDistance} adapted for PPs \cite{hoffman2004multitarget_distance}.
Consider the sets $X=\{x_{1},...,x_{m}\}$ and $Y=\{y_{1},...,y_{n}\}$
as collections of earth piles at $x_{i}$ each with mass $1/m$ and
$y_{j}$ each with mass $1/n$, i.e., the total mass of each collection
is $1$, and suppose that the cost of moving a mass of earth over
a distance is given by the mass times the distance. Then the Wasserstein
distance (\ref{eq:OMAT-dist}) can be considered as the minimum cost
needed to build one collection of earth piles from the other. This
is illustrated in Figs. \ref{fig:Wasserstein_demo} and \ref{fig:EarthMoversDistance},
where the arrows correspond to the optimal movements of the earth
piles. Indeed the Earth Mover's distance has been used to construct
a Support Vector Machine kernel for MI classification in \cite{zhang2007EMD_kernel,amores2013multiple_intance_review}. 

The Wasserstein distance partially addresses the cardinality insensitivity
and reduces the undesirable penalty on the outliers of the Hausdorff
distance \cite{schuhmacher2008_OSPA}, see for example Fig. \ref{fig:Hausdorff_demo}.
However, to the best of our knowledge, it has not been used in MI,
and still has a number of drawbacks.
\begin{itemize}
\item It is still possible for the Wasserstein distance to group together
dissimilar sets while separating similar sets as illustrated in Fig.
\ref{fig:Wasserstein_demo}. Intuitively $X$ and $Z$ are very similar
whereas $X$ and $Y$ are quite dissimilar, but the Wasserstein distance
disagrees, i.e., $d_{\mathtt{W}}^{(2)}(X,Y)\approx1.6<d_{\mathtt{W}}^{(2)}(X,Z)\approx2.3$.
The large Wasserstein distance between $X$ and $Z$ is due to the
moving of earth from the bottom blue pile in Fig. \ref{fig:EarthMoversDistance-b}
over long distances (the two longest blue arrows to red piles in Fig.
\ref{fig:EarthMoversDistance-b}). Note that the elements of $Z$
are not so balanced around the elements of $X$, and thus require
the pile to be moved over long distances. On the other hand the elements
of $Y$ are more balanced around the elements of $X$ thereby requiring
less work and hence a smaller resulting distance. In general, the
Wasserstein distance depends on how well balanced the numbers of points
of $X$ are distributed among the points of $Y$. 
\item Both the Wasserstein and Hausdorff distances are not defined if one
of the sets is empty. However, in PP data, empty PPs are not unusual.
For example, in WiFi log data where each datum (a log record) is a
set of WiFi access point IDs around the scanning device at a given
time, there are instances when there are no WiFi access points leading
to empty observations. In image data where each image is represented
by a set of features describing some objects of interest, images without
any object of interest are represented by empty PPs. 
\end{itemize}

\subsection{OSPA distance\label{subsec:OSPA-distance} }

The Optimal SubPattern Assignment (OSPA) \cite{schuhmacher2008_OSPA}
distance of order $p\geq1$, and cutoff $c>0$, is defined by\vspace{-0mm}
\begin{align}
 & d_{\mathtt{O}}^{(p,c)}(X,Y)=\nonumber \\
 & \hspace{-2mm}\left(\frac{1}{n}\left(\min_{\pi\in\Pi_{n}}\sum_{i=1}^{m}\underline{d}^{(c)}\left(x_{i},y_{\pi(i)}\right)^{p}+c^{p}\left(n-m\right)\right)\right)^{\frac{1}{p}},\label{eq:OSPA-dist}
\end{align}
if $n\geq m>0$ (recall that $m$ and $n$ are the cardinalities of
$X$ and $Y$, respectively), and $d_{\mathtt{O}}^{(p,c)}(X,Y)=d_{\mathtt{O}}^{(p,c)}(Y,X)$
if $m>n>0$, where $\Pi_{n}$ is the set of permutations of $\left\{ 1,2,...,n\right\} $,
$\underline{d}^{(c)}(x,y)=\min\left(c,\underline{d}\left(x,y\right)\right)$.
Further $d_{\mathtt{O}}^{(p,c)}(X,Y)=c$ if one of the set is empty;
and $d_{\mathtt{O}}^{(p,c)}(\emptyset,\emptyset)=0$. The two adjustable
parameters $p$, and $c$, are interpreted as the outlier penalty
and the cardinality sensitivity, respectively.

\vspace{-1mm}
\begin{figure}[tbh]
\begin{centering}
\begin{minipage}[c]{50mm}%
\begin{center}
\includegraphics[width=1\columnwidth]{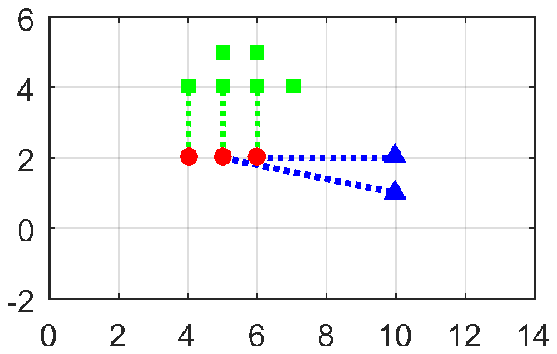}
\par\end{center}%
\end{minipage}\hspace{3mm}%
\begin{minipage}[c]{35mm}%
\begin{center}
\includegraphics[width=1\columnwidth]{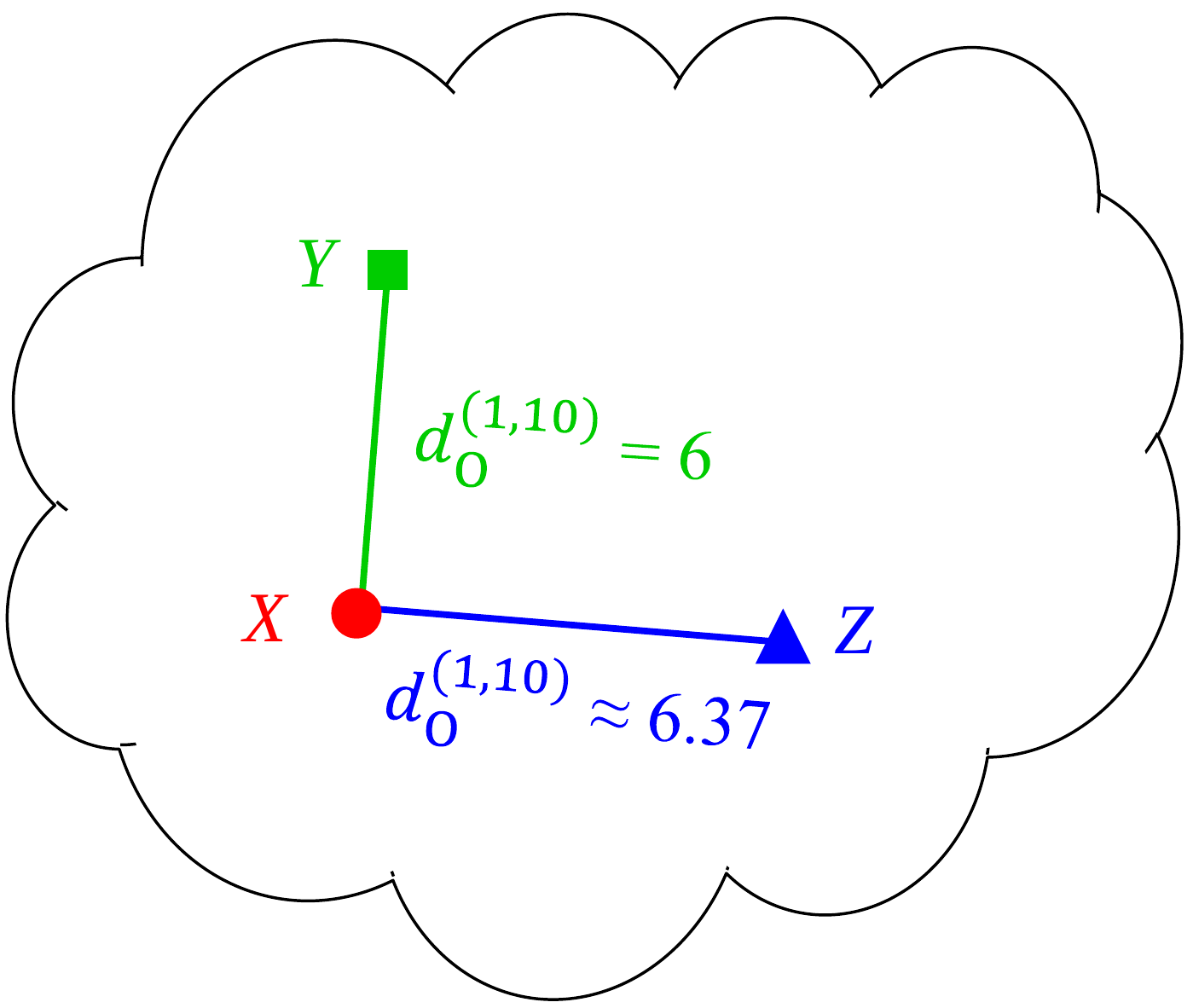}
\par\end{center}%
\end{minipage}
\par\end{centering}
\begin{centering}
\vspace{-0mm}
\caption{\label{fig:OSPA_computing}Computing OSPA distance. Left: Sets $X$
(red ${\color{red}\bullet}$), $Y$ (green {\tiny{}${\normalcolor {\color{green}\blacksquare}}$}),
and $Z$ (blue{\footnotesize{} ${\color{blue}\blacktriangle}$}) in
$\mathbb{R}^{2}$; the dotted lines are optimal assignments between
the elements of $X$ and $Y$, $Z$ respectively. Right: Abstract
impression of the OSPA distances between the sets $X$, $Y$, and
$Z$.}
\par\end{centering}
\vspace{-1mm}
\end{figure}

Assuming $p=1$, to compute (\ref{eq:OSPA-dist}), we assign $m$
elements of $Y$ to the $m$ elements of $X$ so as to minimize the
total adjusted distance $\underline{d}^{(c)}$ (see Fig.~\ref{fig:OSPA_computing}
for illustration). This can be achieved via an optimal assignment
procedure such as Hungarian method. For each of the $\left(n-m\right)$
elements in $Y$ which are not assigned, we set a fixed distance of
$c$. The OSPA distance is simply the average of these $n$ distances
(i.e., $m$ optimal adjusted distances and $\left(n-m\right)$ fixed
distances $c$). Thus, the OSPA distance has a physically intuitive
interpretation as the ``per element'' dissimilarity that incorporates
both features and cardinality \cite{schuhmacher2008_OSPA}. 

\begin{figure}[tbh]
\noindent \begin{centering}
\begin{minipage}[c]{48mm}%
\begin{center}
\includegraphics[width=1\columnwidth]{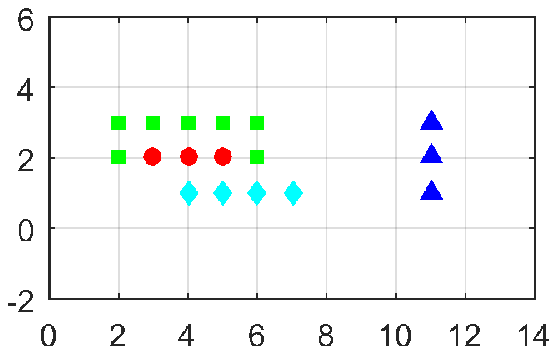}
\par\end{center}%
\end{minipage}\hspace{3mm}%
\begin{minipage}[c]{35mm}%
\begin{center}
\includegraphics[width=1\columnwidth]{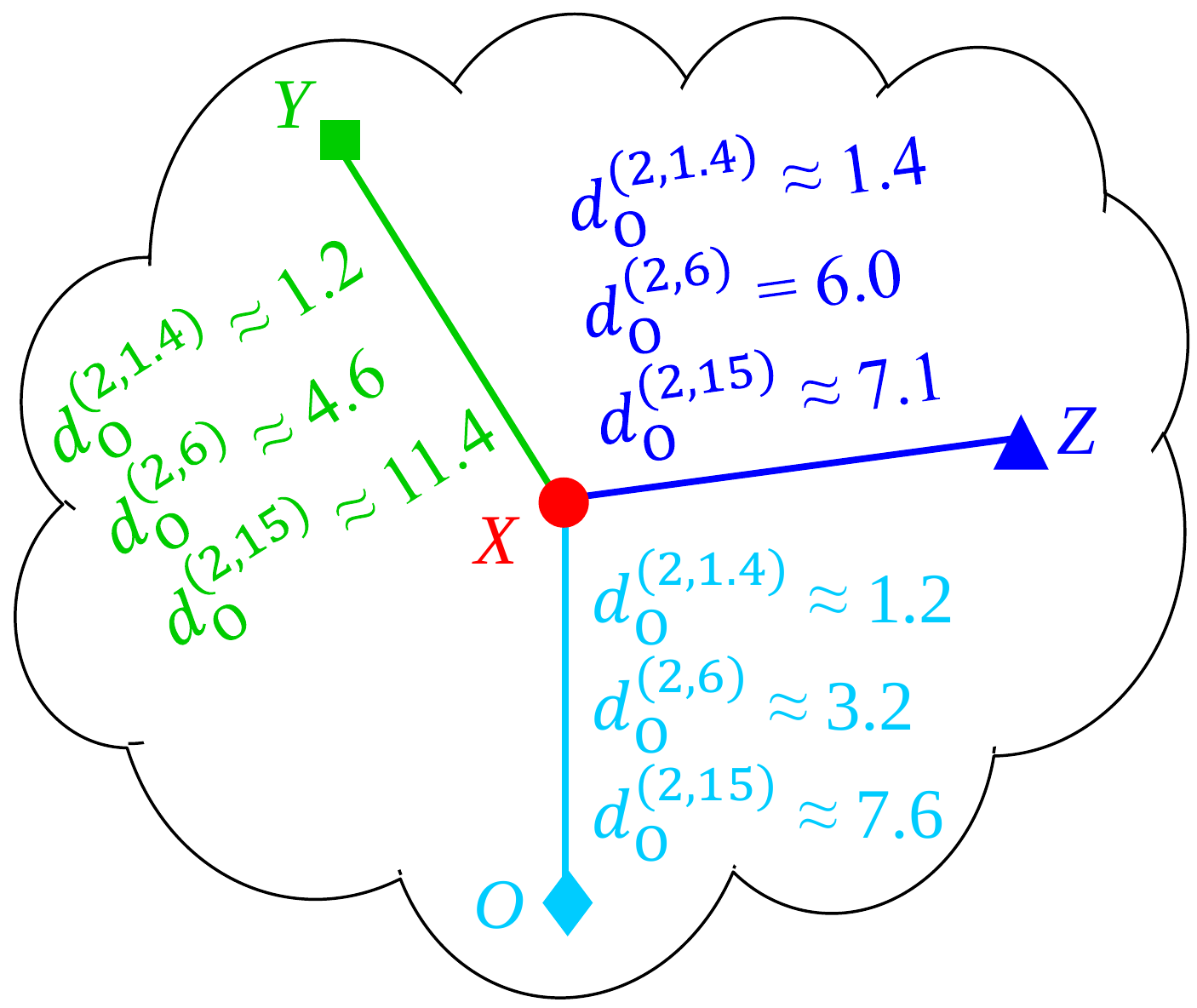}
\par\end{center}%
\end{minipage}
\par\end{centering}
\caption{\label{fig:OSPA_demo} Trade-offs between feature dissimilarity and
cardinality difference in the OSPA distance. Left: Sets $X$ (red
${\color{red}\bullet}$), $Y$ (green {\tiny{}${\normalcolor {\color{green}\blacksquare}}$}),
$Z$ (blue{\footnotesize{} ${\color{blue}\blacktriangle}$}), and
$O$ (cyan{\footnotesize{} ${\color{cyan}{\color{cyan}\blacklozenge}}$})
in $\mathbb{R}^{2}$. Right: Abstract impression of the OSPA distances
between the sets $X$, $Y$, $Z$, and $O$.}
\vspace{-1mm}
\end{figure}

The OSPA distance is a metric with several salient properties that
can address some of the undesirable effects of the Hausdorff and Wasserstein
distances \cite{schuhmacher2008_OSPA}.
\begin{itemize}
\item The OSPA distance penalizes relative differences in cardinality in
an impartial way by introducing an additive component on top of the
average distance in the optimal sub-pattern assignment. The first
term in (\ref{eq:OSPA-dist}) is the dissimilarity in feature while
the second term is the dissimilarity in cardinality. 
\item The OSPA distance is defined for any two PPs. It is equal to $c$
(i.e., maximal) if only one of the two PPs is empty, and zero if both
PPs are empty.
\item \emph{The outlier penalty can be controlled via parameter $p$}. The
larger $p$, the heavier penalty on outliers. Note that the role of
$p$ in OSPA is similar to that for the Wasserstein distance, however,
it is mitigated due to the cutoff $c$. In practice it is common
to use $p=2$.\footnote{For the rest of this paper, we use $p=2$, unless stated.}
\item \emph{The cutoff parameter $c$ controls the trade-offs between feature
dissimilarity and cardinality difference} (see Fig. \ref{fig:OSPA_demo}
for illustration). Indeed, $c$ determines the penalty for cardinality
difference and is also the largest allowable base distance between
constituent elements of any two sets. As a general guide: 1) to\emph{
emphasize feature dissimilarity}, $c$ should be as \emph{small} as
the typical base distance between constituent elements of the PPs
in the given dataset; 2) conversely, to \emph{emphasize cardinality
difference}, $c$ \emph{should be larger than} the maximum base distance
in the given dataset; 3) for a \emph{balanced emphasis} on cardinality
and feature, a \emph{moderate} value of $c$ in between the two aforementioned
values should be chosen. 
\end{itemize}
\vspace{-0mm}

Fig.~\ref{fig:OSPA_demo} shows four PPs $X$, $Y$, $Z$ and $O$,
where: $Y$ has elements that are closest to individual elements of
$X$, but has a larger cardinality; $Z$ has elements far away from
the elements of $X$, but has the same cardinality; while $O$ is
visually most similar to $X$. In this scenario, the typical base
distance between the elements of the PPs is about $1.4$ and the maximum
base distance is about $9.5$. Choosing a small cutoff $c=1.4$ yields
$d_{\mathtt{O}}^{(2,1.4)}(X,Y)<d_{\mathtt{O}}^{(2,1.4)}(X,Z)$, indicating
an emphasis of feature dissimilarity over cardinality difference.
Choosing a large cutoff $c=15$ yields $d_{\mathtt{O}}^{(2,15)}(X,Y)>d_{\mathtt{O}}^{(2,15)}(X,Z)$,
indicating an emphasis of cardinality difference over feature dissimilarity.
Choosing a moderate cutoff $c=6$, makes $O$ closest to $X$, indicating
a balanced emphasis on both feature dissimilarity and cardinality
difference.

We stress that while the OSPA distance offers more flexibility in
design choices and some merits over the other distances, there is
no single distance that works for all applications. 

\section{Clustering of Point Patterns\label{sec:Clustering}}

In general, clustering is an unsupervised learning problem since the
class (or cluster) labels are not provided \cite{Jain1999data_clustering,russell2003artificial}.
The aim of clustering is to partition the data into groups so that
members in a group are similar to each other whilst dissimilar to
observations from other groups~\cite{murphy2012machine}. Clustering
is a fundamental problem in data analysis with a long history dated
back to the 1930s in psychology~\cite{tryon1939cluster}. Comprehensive
surveys on clustering can be found in \cite{Jain1999data_clustering,jain2010clustering50yearsKmeans,xu2005survey_clustering}.

\subsection{Problem Formulation}

In a MI clustering context, the overall goal is to partition a given
PP dataset $\mathcal{D}=\left\{ X_{1},...,X_{N}\right\} \subseteq\mathbb{X}$
into disjoint clusters which minimize the sum of (set) distances between
PPs and their cluster centers, while penalizing the trivial partition
$\mathcal{P}=\left\{ \{X_{1}\},...,\{X_{N}\}\right\} $ (i.e., each
observation is a cluster) that yields zero sum of distances. More
concisely, let $\boldsymbol{\mu}:\mathcal{D}\rightarrow\mathbb{X}$
be a mapping that assigns a cluster center to each PP in $\mathcal{D}$,
i.e., $\boldsymbol{\mu}\left(X\right)$ is the center of the cluster
that $X$ belongs to, then the clustering problem can be stated as
\begin{align}
 & \underset{\boldsymbol{\mu}}{\text{min}}\sum_{X\in\mathcal{D}}d\left(X,\boldsymbol{\mu}(X)\right)+\gamma(X)\delta_{X}[\mu\left(X\right)],\label{eq:clustering1}\\
\text{} & \text{subject to }\boldsymbol{\mu}\left(C\right)=C,\forall C\in\boldsymbol{\mu}(\mathcal{D}),\label{eq:clustering2}
\end{align}
where $\delta_{A}[B]=1\text{ if }A=B,\text{ and is }0\text{ otherwise}$,
$\gamma:\mathcal{D}\rightarrow[0,\infty)$ is a user chosen penalty
function that imposes a penalty for the selection of an observation
$X$ as its own cluster centre, and hence penalizes the identity map
$\boldsymbol{\mu}:X\mapsto X$ as a solution. 

\emph{Remark:} The mapping $\boldsymbol{\mu}$ provides a partitioning
$\mathcal{P}=\left\{ \mathcal{P}_{1},...,\mathcal{P}_{\mid\boldsymbol{\mu}\left(\mathcal{D}\right)\mid}\right\} $
of the dataset $\mathcal{D}$, where $\mathcal{P}_{k}=\left\{ X\in\mathcal{D}:\boldsymbol{\mu}\left(X\right)=C_{k}\right\} $
is the $k^{\text{th}}$ cluster and $\boldsymbol{\mu}\left(\mathcal{D}\right)=\left\{ C_{1},...,C{}_{\mid\boldsymbol{\mu}\left(\mathcal{D}\right)\mid}\right\} $
is the set of cluster centers or centroids. The constraints ensure
that if a PP $C\in\mathcal{D}$ is a cluster centre, then $C$ must
belong to the cluster with centre $C$. The user defined penalty $\gamma(X)$
can also be interpreted in terms of the preference for datum $X$
to be a centroid: the smaller $\gamma(X)$ is, the higher we prefer
$X$ to be a centroid.

Note that the cluster center $\boldsymbol{\mu}\left(X\right)$ of
a datum $X$ can be either defined as the mean (or more generally
the Fréchet mean) of the observations in its group (e.g., $k$-means)
or chosen among observations in the dataset, i.e., $\boldsymbol{\mu}:\mathcal{D}\rightarrow\mathcal{D}$
(e.g., $k$-medoids). In general, the Fréchet mean of a collection
of PPs is computationally intractable \cite{baum2015ospa_clustering}
and a better strategy is to select the centroids from the dataset.
Such centroids, also known as `exemplars'~\cite{frey2007_APclustering},
can be efficiently computed as well as serving as real prototypes
for the data. 

To the best of our knowledge, BAMIC \cite{zhang2009MIClustering}
is the only exemplar-based clustering algorithm for PPs using a set
distance (Hausdorff) as a measure of dissimilarity. The Hausdorff
distance, used by BAMIC, has several undesirable properties as discussed
in section \ref{subsec:Hausdorff-distance}. Moreover, since BAMIC
is based on the $k$-medoids algorithm, it requires the number of
clusters as an input, which is not always available in practice. Determining
the correct number of clusters is one of the most challenging aspects
of clustering~\cite{jain2010clustering50yearsKmeans}. While it is
possible to perform model selection via cross-validation for different
number of clusters to decide on the best one, this process incurs
substantial computational cost. In addition, it is mathematically
more principled to jointly determine the number of clusters and their
centers. 

In this work, we propose a versatile MI clustering algorithm using
the AP algorithm \cite{frey2007_APclustering} with the OSPA distance
as a dissimilarity measure. For the sake of performance comparison,
we also include the Hausdorff and Wasserstein distances as baselines.
Using message passing, AP provides good approximate solutions to problem
(\ref{eq:clustering1})-(\ref{eq:clustering2}) \cite{dueck2007non_metric_AP,frey2007_APclustering},
thereby determining the number of clusters automatically from the
data (see details in section \ref{subsec:AP-clustering}). Compared
to $k$-medoids (used in BAMIC), AP can find clusters faster with
considerably lower error~\cite{frey2007_APclustering} and does not
require random initialization of cluster centers (since AP first considers
all observations as exemplars). In addition, the OSPA distance does
not suffer from the undesirable effects as the Hausdorff distance
used in BAMIC, as well as being more flexible (section \ref{subsec:OSPA-distance}). 

\subsection{AP clustering with set distances\label{subsec:AP-clustering}}

The AP algorithm has been widely used in several applications due
to its ability to automatically infer the number of clusters and fast
execution time. 

The AP algorithm uses the similarity values between all pairs of observations
in the data set $\mathcal{D}=\{X_{1},...,X_{N}\}$ and user defined
exemplar\emph{ preferences}, as input and returns the `best' set of
exemplars. The similarity values of interest in this work are the
negatives of the OSPA distances between the PPs in $\mathcal{D}$.
The preference value for a datum $X_{n}$ is the negative of the penalty,
i.e., $-\gamma(X_{n})$, the larger its preference, the more likely
that $X_{n}$ is an exemplar. In AP, the exemplar for an observation
$X_{n}$ (which could be $X_{n}$ itself or another observation) is
represented by a variable $c_{n}$, where $c_{n}=k$ means that $X_{k}$
is the exemplar for $X_{n}$. 

Note that a configuration $(c_{1},\ldots,c_{N})$ provides an equivalent
representation of the decision variable $\boldsymbol{\mu}:\mathcal{D}\rightarrow\mathcal{D}$
in problem (\ref{eq:clustering1})-(\ref{eq:clustering2}) by defining
$\boldsymbol{\mu}\left(X_{n}\right)=X_{k}$ iff $c_{n}=k$. Treating
each $c_{n}$ as a random variable, a factor graph with nodes $c_{1},\ldots,c_{N}$
can be constructed by encoding into the functional potentials the
similarities between pairs of observations, the preferences for each
observation, as well as constraints that ensure valid cluster configurations.
Constraint (\ref{eq:clustering2}) means that in a valid configuration
$(c_{1},\ldots,c_{N})$, $c_{c_{n}}=c_{n}$, i.e., if $X_{k}$ is
an exemplar for any observation, then the exemplar of $X_{k}$ is
$X_{k}$. This constraint can be enforced by setting the potential
of any configuration $(c_{1},\ldots,c_{N})$ with $c_{c_{n}}\neq c_{n}$
to $-\infty$. Ideally, performing max-sum message-passing yields
a configuration that maximizes the sum of all potentials in this factor
graph, and hence a solution to the clustering problem (\ref{eq:clustering1})-(\ref{eq:clustering2}).
AP is an efficient approximate max-sum message-passing algorithm using
a protocol originally derived from loopy propagation on factor graphs
\cite{frey2007_APclustering}.\footnote{An equivalent binary graphical model representation for AP was later
proposed in~\cite{givoni2009binary_AP}. Instead of creating a latent
node for each individual observation as in \cite{frey2007_APclustering},
a binary node $b_{n,k}$ is created for each pair $\left(X_{n},X_{k}\right)$
and $b_{n,k}=1$ if $X_{k}$ is an exemplar for $X_{n}$. Message-passing
on this new factor graph representation yields the same solution.} Further details on the AP algorithm can be found in \cite{frey2007_APclustering,dueck2007non_metric_AP}.
In what follows, we discuss specific details for the clustering of
PP data summarized in Algorithm~\ref{alg:AP_algorithm}. 

\begin{algorithm}[tbh]
\Indentp{-1.6mm}

\SetKwInOut{Input}{Input}
\Input{PP dataset $\mathcal{D}=\left\{ X_{1},...,X_{N}\right\} $,\\

Stopping threshold $\theta$.}

\SetKwInOut{Output}{Output}
\Output{Cluster labels $c_{n}$ where $n\in\{1,...,N\}$.}


\For{$n,k\in\{1,...,N\}$ }{

\tcp{{\small{}Compute similarities}}

$s(n,k)=-d_{\mathtt{O}}\left(X_{n},X_{k}\right)$;

\tcp{{\small{}Assign preferences}}

$s(n,n)=-\gamma(X_{n})$;

\tcp{{\small{}Initialize messages}}

$r(n,k)=0$; $a(n,k)=0$;}

\Repeat{change of any $r(\cdot,\cdot)$ or $a(\cdot,\cdot)$ $<\theta$}{

\For{$n,k\in\{1,...,N\}$}{

\tcp{{\small{}Update responsibilities}}\vspace{-5mm}
\begin{equation}
\hspace{-5mm}r\left(n,k\right)=s\left(n,k\right)-\max_{k'\neq k}\{a\left(n,k'\right)+s\left(n,k'\right)\};\hspace{-1mm}\label{eq:AP_responsibility}
\end{equation}
\vspace{-3mm}

\tcp{{\small{}Update availabilities}}\vspace{-5mm}

\begin{align}
\hspace{-3.7mm}a\left(n,k\right) & =\min\{0,r\left(k,k\right)\}\hspace{-0.6mm}+\hspace{-4mm}\sum_{n'\notin\left\{ n,k\right\} }\hspace{-4mm}\max\{0,r\left(n',k\right)\};\label{eq:AP_availability}
\end{align}
\vspace{-4mm}
\begin{align}
\hspace{-11mm}a\left(k,k\right) & =\sum_{n'\ne k}\max\{0,r\left(n',k\right)\};\hspace{14mm}\label{eq:AP_self_availability}
\end{align}

\vspace{-3mm}
}

}

\tcp{{\small{}Return cluster assignments}}

\For{$n\in\{1,...,N\}$}{

$c_{n}=\underset{k}{\text{argmax}}\,\left(r(n,k)+a(n,k)\right)$;}

\caption{\label{alg:AP_algorithm}Clustering of PP data using set distances.
See text for values of $\gamma(X_{n})$.}
\end{algorithm}

The algorithm starts by computing all pairwise similarities input
for AP: $s(n,k)=-d_{\mathtt{O}}(X_{n},X_{k})$, and preferences $s\left(k,k\right)=-\gamma(X_{k})$.
A common practice is to give all observations the same preference,
e.g., the median of the similarities (which results in a moderate
number of clusters) or the minimum of the similarities (which results
in a small number of clusters) \cite{frey2007_APclustering}.

The AP algorithm passes two types of messages. The \emph{responsibility}
$r(n,k)$, defined in (\ref{eq:AP_responsibility}) indicating how
well $X_{n}$ trusts $X_{k}$ as its exemplar, is sent from observation
$X_{n}$ to its candidate exemplar $X_{k}$. Then, the \emph{availability}
$a(n,k)$, defined in (\ref{eq:AP_availability}) reflecting the accumulated
evidence for $X_{k}$ to be an exemplar for $X_{n}$, is sent from
a candidate exemplar $X_{k}$ to $X_{n}$. Note from (\ref{eq:AP_responsibility})-(\ref{eq:AP_availability})
that the responsibility $r\left(n,k\right)$ is calculated from availability
values that $X_{n}$ receives from its potential exemplar, whereas
the availability $a\left(n,k\right)$ is updated using the `support'
from observations that consider $X_{k}$ as their candidate exemplar.
Note from (\ref{eq:AP_availability}) that when an observation is
assigned to exemplars other than itself, its availability falls below
zero. Such negative availabilities in turn decrease the effect of
input similarities $s\left(n,k'\right)$ in (\ref{eq:AP_responsibility}),
thereby eliminating the corresponding PPs from the set of potential
exemplars.

The loopy propagation is usually terminated when changes in the messages
fall below a threshold (see Algorithm \ref{alg:AP_algorithm}), or
when the cluster assignments stay constant for some iterations, or
when number of iterations reaches a given value~\cite{frey2007_APclustering}.
The cluster label $c_{n}$ is the value of $k$ that maximizes the
sum $r(n,k)+a(n,k)$~\cite{frey2007_APclustering}. 

\subsection{Experiments\label{subsec:AP-experiments}}

In this section we evaluate the performance of the proposed AP-based
clustering algorithm on both simulated and real PP data. In particular,
we compare the clustering performance amongst the Hausdorff, Wasserstein
and OSPA  distances. Note that BAMIC (which uses the $k$-medoids
algorithm instead of AP) can be treated as AP clustering with the
Hausdorff distance. 

Since the result of the AP algorithm depends on the choice of exemplar
preferences, we first empirically select the exemplar preference that
yields the best performance in terms of number of clusters for each
distance, and then benchmark the best case performance of one distance
against the others. The relevant performance indicators are: Purity
(Pu), Normalized mutual information (NMI), Rand index (RI), F\textsubscript{1}
score (F1) \cite{manning2008info_retrieval}.

\subsubsection{Clustering with simulated data\label{subsec:AP-sim}}

\begin{figure*}[tbh]
\begin{centering}
\medskip{}
\par\end{centering}
\begin{centering}
\subfloat[\label{fig:AP_sim_sepa_feat}Dataset (i)]{\begin{centering}
\includegraphics[height=4.6cm]{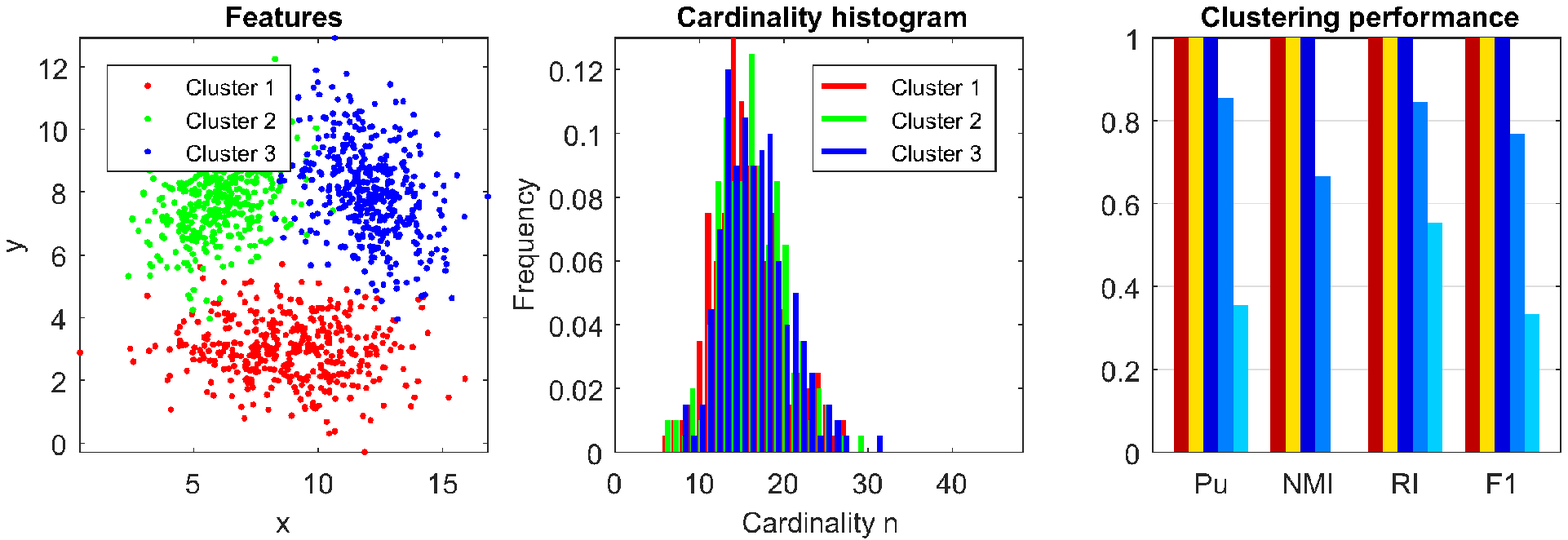}~\includegraphics[height=4.2cm]{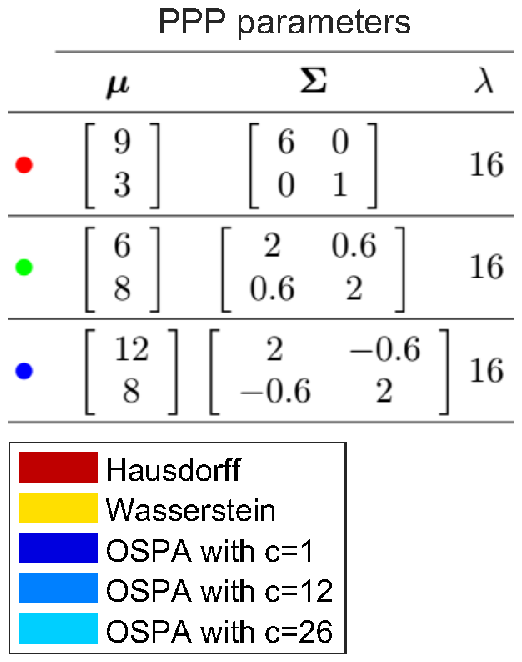}
\par\end{centering}
\centering{}\vspace{-4mm}
}
\par\end{centering}
\begin{centering}
\vspace{-3mm}
\subfloat[\label{fig:AP_sim_sepa_card}Dataset (ii)]{\begin{centering}
\includegraphics[height=4.4cm]{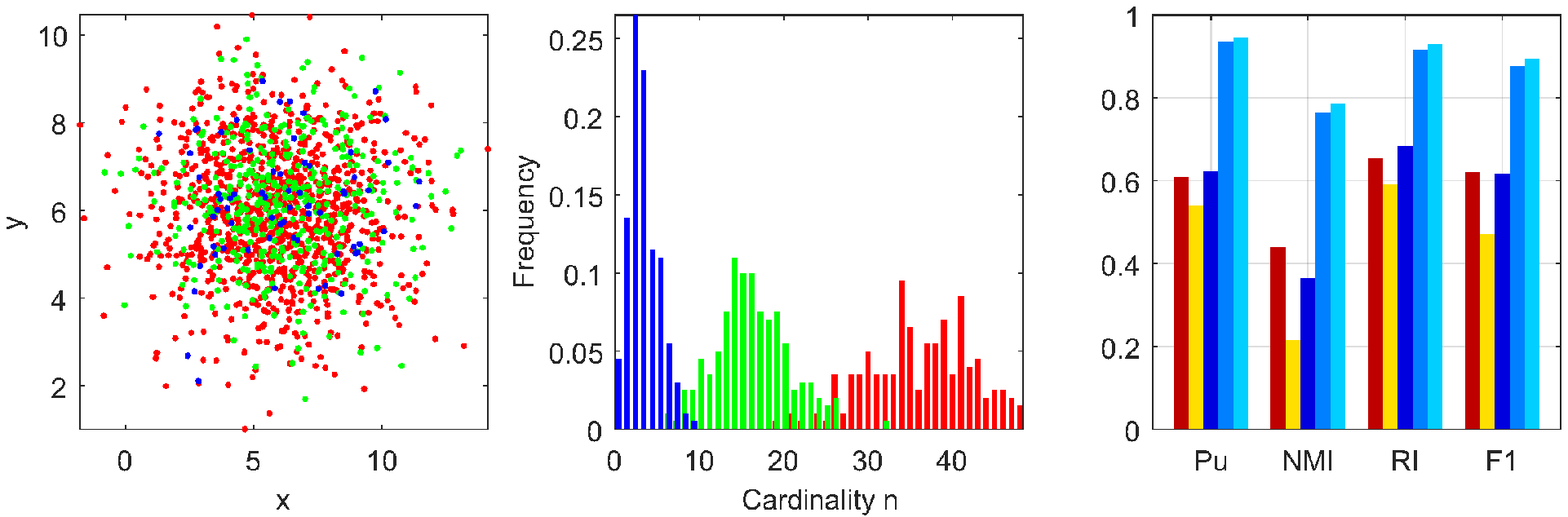}~\includegraphics[height=4.2cm]{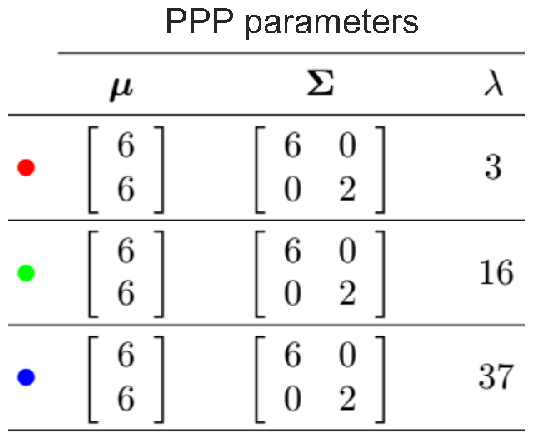}
\par\end{centering}
\centering{}\vspace{-4mm}
}\vspace{-3mm}
\par\end{centering}
\begin{centering}
\subfloat[\label{fig:AP_sim_mix}Dataset (iii)]{\begin{centering}
\includegraphics[height=4.4cm]{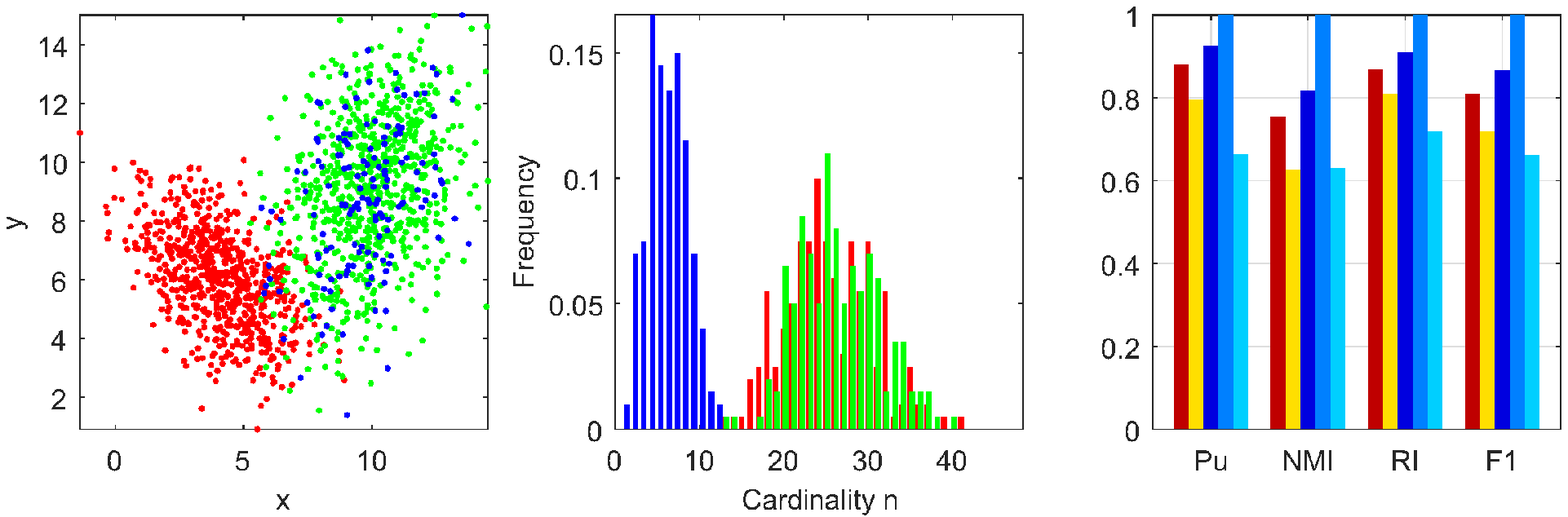}~\includegraphics[height=4.2cm]{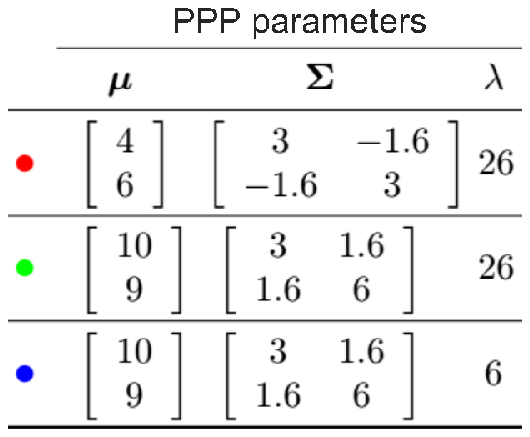}
\par\end{centering}
\centering{}\vspace{-2mm}
}
\par\end{centering}
\vspace{-0mm}
\caption{\label{fig:AP-sim}Clustering performance for three diverse scenarios.
Dataset (i): well-separated in feature but overlapping in cardinality;
dataset (ii): well-separated in cardinality but overlapping in feature;
data set (iii): a mix of (i) and (ii). Performance indicators: Purity
(Pu); Normalized mutual information (NMI); Rand index (RI); and F\protect\textsubscript{1}
score (F1).}

\vspace{-3mm}
\end{figure*}

In this experiment, we consider three simulated datasets. Each dataset
consists of 3 clusters, each cluster consists of 200 PPs generated
from a Poisson point process (PPP) with a 2-D Gaussian intensity.\footnote{This dataset is similar to that of \cite{vo2016model-based_PP}. In
fact, they simulated by the same mechanism. } In brief, a PP is sampled from a PPP with Gaussian intensity parameterized
by $(\lambda,\boldsymbol{\mu},\boldsymbol{\Sigma})$, by first sampling
the number of points from a Poisson distribution with rate $\lambda$,
and then sampling the corresponding number points independently from
the Gaussian with mean and covariance $(\boldsymbol{\mu},\boldsymbol{\Sigma})$.
The parameters for the PPPs used in this experiment are shown in Fig.
\ref{fig:AP-sim}. Three diverse scenarios are considered: in dataset
(i) features of the PPs from each cluster are well separated from
those of the other clusters, but their cardinalities significantly
overlap (see Fig. \ref{fig:AP_sim_sepa_feat}); in data set (ii) cardinalities
of the PPs from each cluster are well separated from those of the
other clusters, but their features significantly overlap (see Fig.
\ref{fig:AP_sim_sepa_card}); dataset (iii) is a mix of (i) and (ii)
(see Fig. \ref{fig:AP_sim_mix}). 

Three different cutoff values for the OSPA distance are experimented:
$c=1$ (small); $c=12$ (moderate); and $c=26$ (large). Note that
$c=1$ is a typical value of the intra-PP base distance (i.e., base
distance between the features within the PPs in the dataset), $c=26$
is an estimate of the maximum intra-PP base distance, and $c=12$
is a moderate value of the intra-PP base distance. 

In dataset (i) (Fig. \ref{fig:AP_sim_sepa_feat}), the Hausdorff,
Wasserstein and OSPA distances with small and moderate cutoffs show
good performance. The OSPA distance with a large cutoff tends to emphasize
the cardinality dissimilarities (which are negligible in this scenario)
over feature dissimilarities (see subsection \ref{subsec:OSPA-distance})
leading to poor clustering performance. 

In dataset (ii) (Fig. \ref{fig:AP_sim_sepa_card}), where cardinality
difference is the main discriminative information, the Hausdorff and
Wasserstein distances perform poorly since they are unable to capture
cardinality dissimilarities between the PPs. The OSPA distance with
a small cutoff tends to emphasize feature dissimilarities (which are
negligible in this scenario) over cardinality dissimilarities (see
section \ref{subsec:OSPA-distance}) leading to poor performance.
On the other hand the OSPA distance with moderate and large cutoffs
perform better since they can appropriately capture cardinality dissimilarities. 

In dataset (iii) (Fig. \ref{fig:AP_sim_mix}), the results again confirm
the discussions above. The OSPA distance with moderate cutoff provides
a balanced emphasis on both feature and cardinality dissimilarities,
yielding the best performance. 

\subsubsection{Clustering with the Texture dataset\label{subsec:AP-Texture}}

\begin{figure}[tbh]
\begin{centering}
\vspace{-0mm}
\includegraphics[width=26mm]{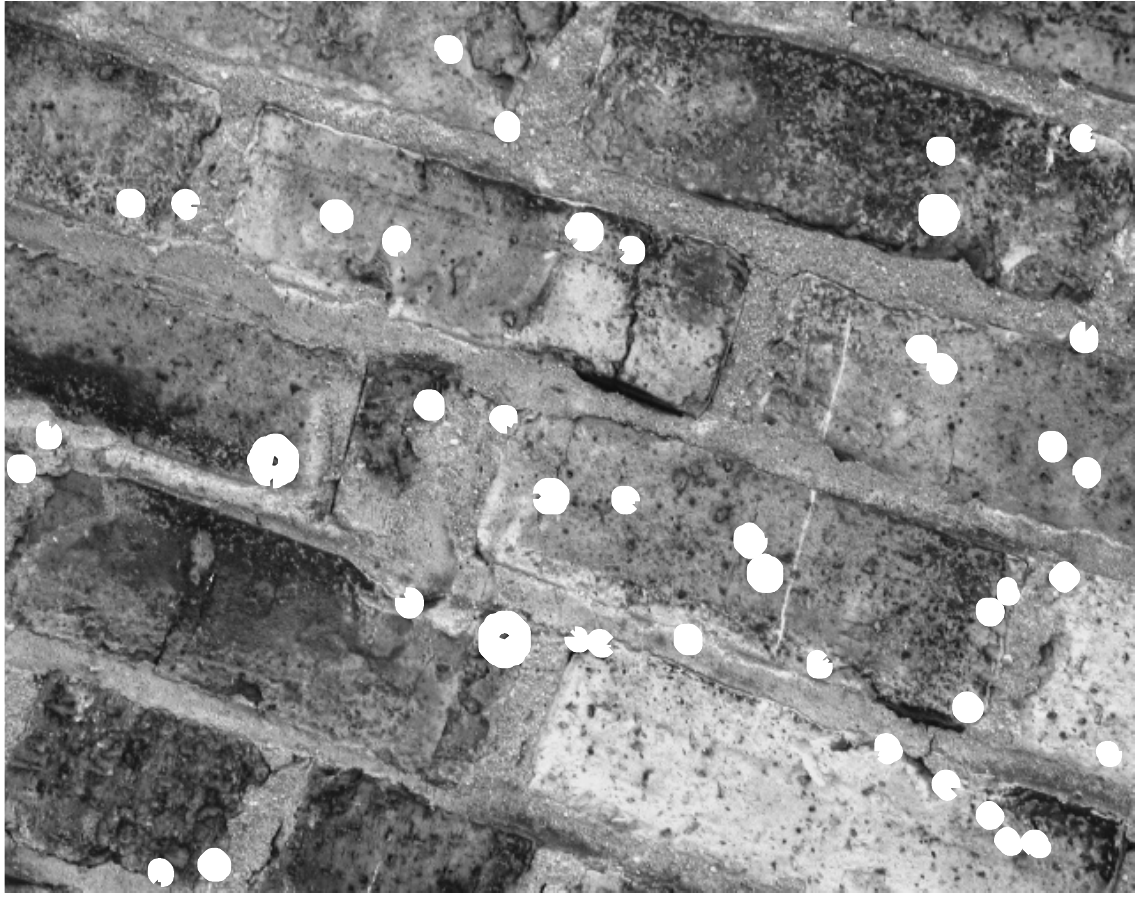}~~\includegraphics[width=26mm]{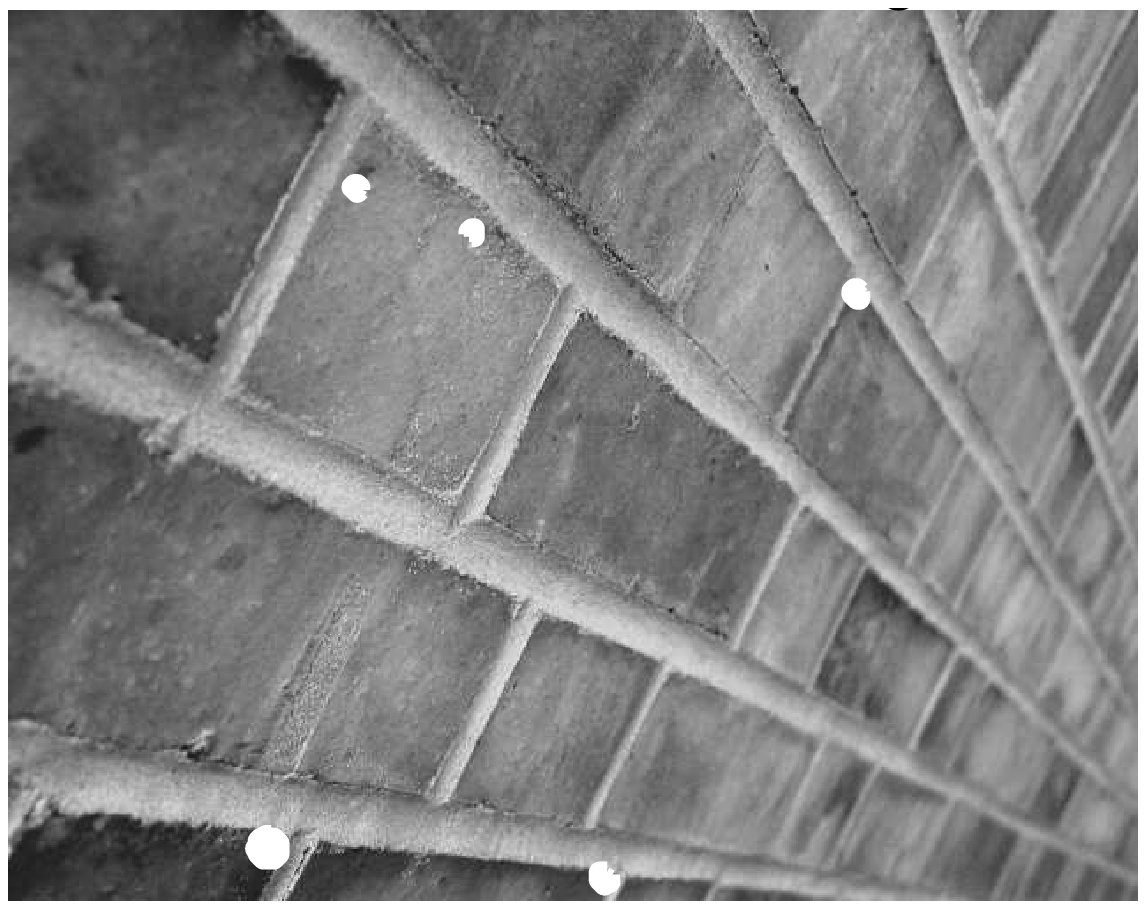}~~\includegraphics[width=26mm]{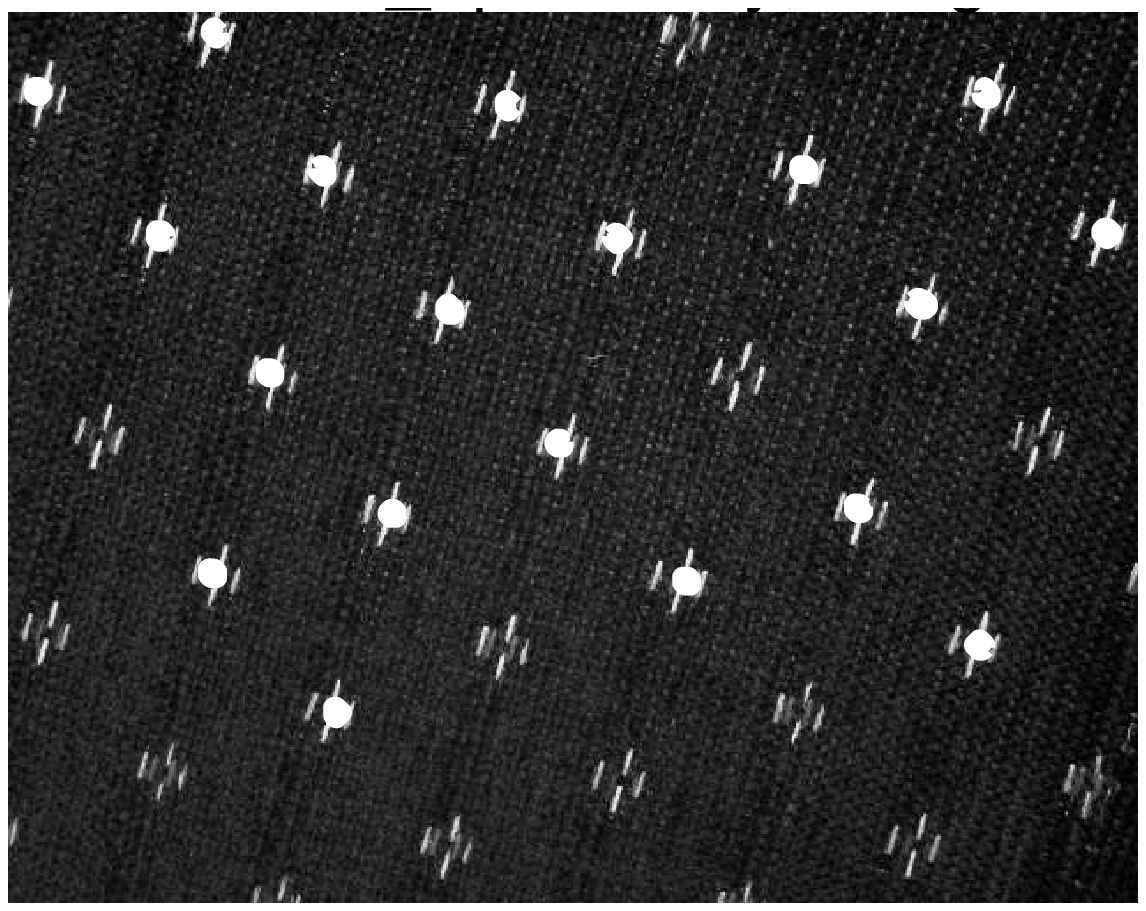}
\par\end{centering}
\begin{centering}
\vspace{-0mm}
\par\end{centering}
\caption{\label{fig:Texture_examples}Example images from the Texture dataset.
From left to right: image from class ``T14 brick1'', ``T15 brick2'',
and ``T20 upholstery''. Circles mark detected SIFT keypoints.}

\vspace{-1mm}
\end{figure}

This experiment involves clustering images from the classes ``T14
brick1'', ``T15 brick2'', and ``T20 upholstery'' of the Texture
images dataset \cite{textureDataset}. Each class consists of 40 images,
with some examples shown in Fig. \ref{fig:Texture_examples}. Each
image is compressed into a PP of 2-D features by first applying the
SIFT algorithm (using the VLFeat library \cite{vlfeatLib}) to produce
a PP of 128-D SIFT features, which is then further compressed into
a 2-D PP by Principal Component Analysis (PCA). Fig. \ref{fig:Texture_data_results}
shows the superposition of the 2-D PPs from the three classes along
with their cardinality histograms. 

Fig. \ref{fig:Texture_data_results} shows that the OSPA distances
(especially with $c=20$) outperform the Hausdorff and Wasserstein
distances, since it can incorporate both feature and cardinality information.
The poor performance of the Hausdorff and Wasserstein distances is
due to the significant overlap in the features and their inability
to measure cardinality dissimilarities in the data.

\subsubsection{Clustering with the StudentLife dataset\label{subsec:AP_studentlife}}

\begin{figure*}[tbh]
\begin{centering}
\vspace{0mm}
\par\end{centering}
\begin{centering}
\includegraphics[width=137mm]{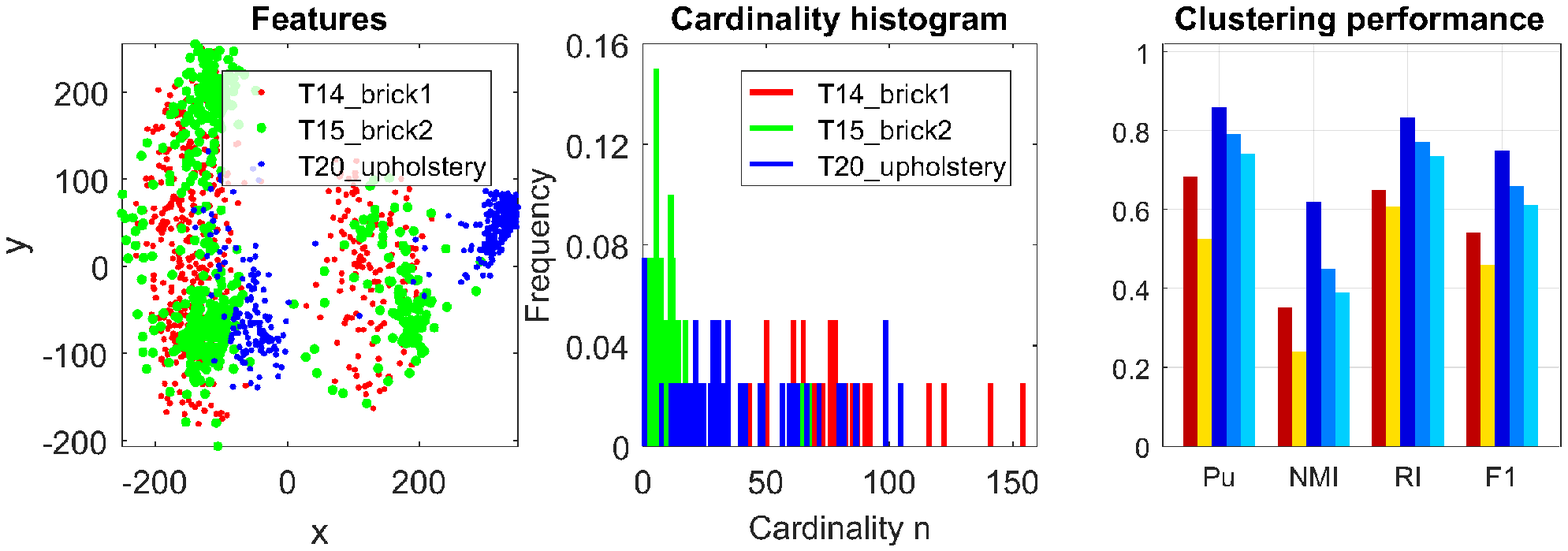}\includegraphics[width=19mm]{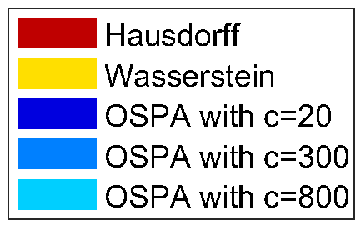}
\par\end{centering}
\vspace{-0mm}
\caption{\label{fig:Texture_data_results}PP data from images of classes ``T14
brick1'', ``T15 brick2'', and ``T20 upholstery'' of the Texture
dataset, and clustering performance for various distances. }

\begin{centering}
\vspace{2mm}
\par\end{centering}
\begin{centering}
\includegraphics[width=137mm]{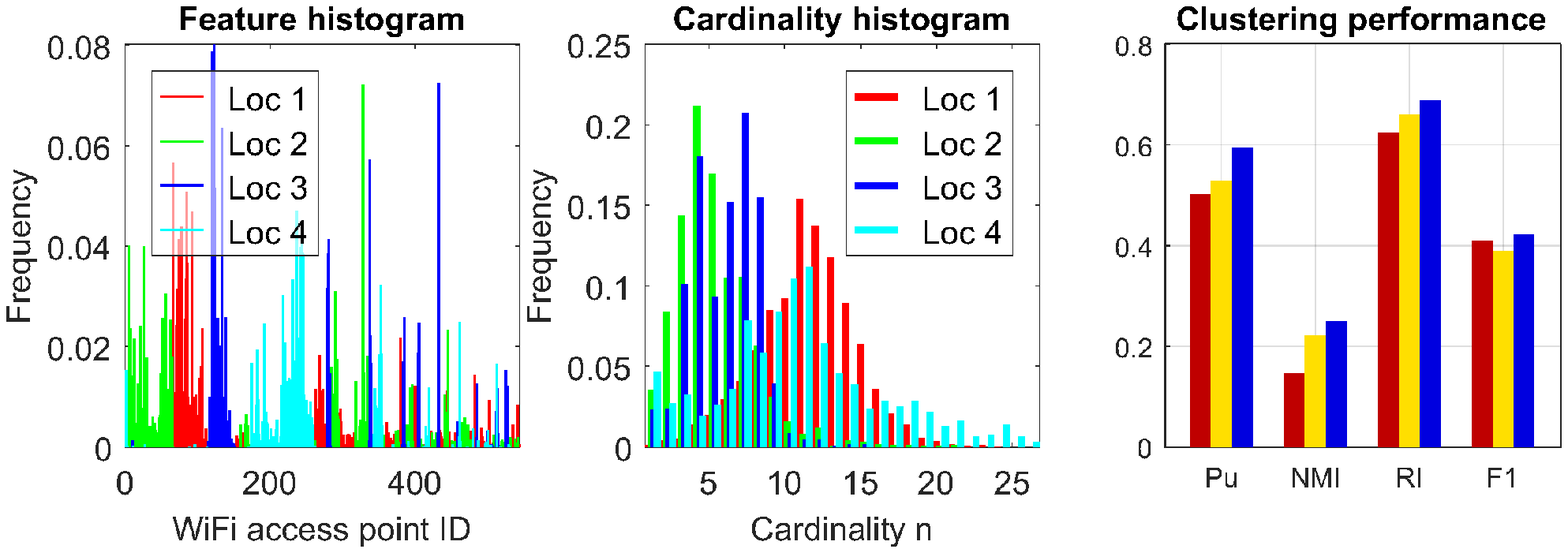}\includegraphics[width=19mm]{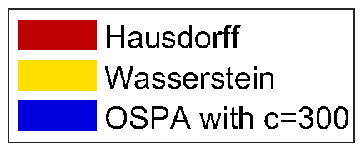}
\par\end{centering}
\vspace{-0mm}
\caption{\label{fig:StudentLife_result_AP}PP data from the StudentLife dataset
and clustering performance for various distances. }

\vspace{-1mm}
\end{figure*}

This experiment involves WiFi scan data from the StudentLife dataset
\cite{studentlife_dataset} collected from smartphones carried by
students at Dartmouth College. At every preset interval, the phone
automatically scans for surrounding WiFi access points and records
detected ones. Therefore, each observation is a PP of WiFi access
point IDs (called \emph{WiFi IDs}). 

The logs of these WiFi scans can be used to infer the history of visited
places since scans containing similar PPs of WiFi IDs are normally
recorded at close-by locations. Thus, estimating the visited locations
from WiFi IDs PPs can be formulated as a clustering problem, where
each cluster represents a visited location. In the StudentLife data
collection, the location of each scan (at the building level) is retrieved
by mapping the detected WiFi IDs to the WiFi deployment information
provided by Dartmouth Network Services. However, this deployment information
is highly protected and is not available to the general public. 

In this experiment data from a random participant is pre-processed
so as to keep only WiFi IDs appearing at least 10 times (544 such
WiFi IDs). Further, only 4 locations that received more observations
than the number of WiFi IDs are considered. Fig.~\ref{fig:StudentLife_result_AP}
shows the frequency histograms of WiFi IDs and cardinality histogram
of the observations collected from 4 considered locations.

For performance assessment, we use the locations provided in the dataset
as ground-truth. Observe from Fig. \ref{fig:StudentLife_result_AP}
that the OSPA distance (other cutoff values have similar performance
to that with $c=300$ and are not shown) performs better than both
the Hausdorff and Wasserstein distances. However, the improvement
is not drastic since there are substantial overlaps in both features
and cardinalities between different clusters.

\vspace{-0mm}

\section{Classification of Point Patterns\label{sec:Classification}}

Classification is the supervised learning task of assigning a class
label $\ell\in\left\{ 1,\ldots,N_{\mathrm{class}}\right\} $ to each
input observation $X$ \cite{bishop2006pattern}. Unlike its unsupervised
counterpart, i.e., clustering (section \ref{sec:Clustering}), classification
relies on training data, which are fully-observed input-output pairs
$\mathcal{D}_{\mathrm{train}}=\{(X_{n},\ell_{n})\}_{n=1}^{N_{\mathrm{train}}}$
\cite{murphy2012machine}. Classification is arguably the most widely
used form of supervised machine learning, spanning various fields
of study \cite{murphy2012machine,witten2005data_mining}.

The classification problem can be approached with or without knowledge
of the underlying data model \cite{cover1967nearest_neighbor_classifier}.
In this paper, we focus on the so-called non-parametric classifiers,
which do not require knowledge of the data model. Among non-parametric
classifiers such as Support Vector Machine (a binary classifier) \cite{boser1992training_svm,cortes1995SVM},
Parzen window \cite{duda2001pattern}, $k$-Nearest Neighbors ($k$-NN)
\cite{cover1967nearest_neighbor_classifier,keller1985fuzzy_kNN},
$k$-NN is more suited to PP data classification using set distances. 

The $k$-NN classifier has two phases: training and classifying. Contrary
to eager learning algorithms in which a model is learned from training
data in the training phase, the $k$-NN algorithm delays most of its
computational effort to the classifying (or test) phase. In the training
phase, the only task is storing class labels of the training observations.
In the test phase, when a new observation is passed to query its label,
the algorithm determines its $k$ nearest observations, with respect
to some distance, in the training set. The queried observation is
then assigned the most popular label among its $k$ nearest neighbours.

\subsection{$k$-NN classification with set distances\label{subsec:k-NN-classification-with-set}}

In MI learning, PP classifiers based on the $k$-NN algorithm using
set distances such as Hausdorff \cite{huttenlocher1993tracking_Hausdorff},
Chamfer \cite{gavrila1999Chamfer_distance}, and Earth Mover's \cite{zhang2007EMD_kernel,rubner1998EarthMoversDistance}
have been proposed. However, the OSPA distance is more versatile and
better at capturing feature and cardinality dissimilarities between
PPs. Hence, the OSPA distance would be more effective with the $k$-NN
algorithm for PP classification.

Unlike existing $k$-NN classification that only stores the class
labels in the training phase, our proposed approach exploits training
data to learn a suitable dissimilarity measure. Since the fully observed
training data can be used to assess whether the set distance agrees
with the notion of similarity/dissimilarity of the application under
consideration, in principle, a suitable distance can be learned. A
simple approach is to perform cross-validation on the training data
for a range of distances and select the best. Intuitively, a suitable
distance entails small dissimilarities between observations in the
same class, but large dissimilarities between observations from different
classes. Hence, for a given training dataset, we seek a distance (or
its parameterization) that minimizes the ratio of inter-class dissimilarity
to intra-class dissimilarity. In general, learning an arbitrary distance
from training data is numerically intractable. However, it is possible
to learn low dimensional parameters such as the cut-off parameter
in the OSPA distance. 

\textit{\emph{The OSPA distance provides the capability for adapting
the weighing between feature dissimilarity and cardinality dissimilarit}}y
via the cut-off parameter $c$. While the right balance between \textit{\emph{feature
and cardinality dissimilarities}} varies from one application to another,
it can be learned from the fully observed training data via cross-validation.
However, cross-validation is not suitable for small training datasets.
In the following, we describe an alternative approach that also accommodates
small datasets. 

Let $\bar{d}_{\mathtt{O}}^{(p,c)}(X,C)$ denote the average OSPA distance,
with cut-off $c$, from a PP $X$ to its $k$ nearest neighbours in
a collection $C$ (of PPs), and let $C_{\ell}$ denote the class of
PP observations with class label $\ell$ in the training set. Then
the inter-class dissimilarity for $C_{\ell}$ is defined by $\hat{D}^{(p,c)}(C_{\ell})=\max_{X\in C_{\ell}}d_{\mathtt{O}}^{(p,c)}(X,C_{\ell})$
while its intra-class dissimilarity is defined by $\check{D}^{(p,c)}(C_{\ell})=\min_{j\neq\ell}\min_{X\in C_{\ell}}d_{\mathtt{O}}^{(p,c)}(X,C_{j})$.
To enforce small inter-class dissimilarity and large intra-class dissimilarity,
we seek cut-off parameters that minimize the worst-case (over the
training data set) ratio of inter-class dissimilarity to intra-class
dissimilarity 
\[
\rho(c)=\max_{\ell}\left(\frac{\hat{D}_{\mathtt{O}}^{(p,c)}(C_{\ell})}{\check{D}_{\mathtt{O}}^{(p,c)}(C_{\ell})}\right)
\]
The operations max, min in the definition of $\rho$ can be replaced
by averaging or a combination thereof. For large training datasets
averaging is preferable. 

\subsection{Experiments\label{subsec:k-NN-experiments}}

In the following experiments, we benchmark the classification performance
of the OSPA distance against the Hausdorff\footnote{and hence the Chamfer ``distance'', see subsection \ref{subsec:Hausdorff-distance}.}
and Wasserstein\footnote{and hence the Earth Mover's distance, see subsection \ref{subsec:Wasserstein-distance}.}
 distances on both simulated and real data. Since the performance
depends on the choice of $k$ (the number of nearest neighbours),
we ran our experiments for each $k\in\{1,...,10\}$ and benchmark
the best case performance of one distance against the others. 

\subsubsection{Classification of simulated data\label{subsec:k-NN-sim}}

\begin{figure}[tbh]
\begin{centering}
\vspace{-1mm}
\par\end{centering}
\begin{centering}
\includegraphics[width=90mm]{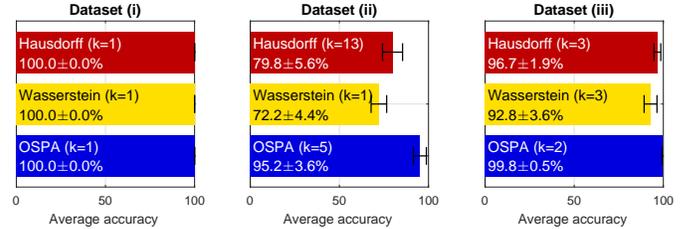}
\par\end{centering}
\vspace{-0mm}
\caption{\label{fig:Sim_data_classification}Classification performance on
simulated data for various distances ($k$ is the number of nearest
neighbours used). The error-bars represent standard deviations of
the accuracies. }

\vspace{-1mm}
\end{figure}

This experiment examines the classification performance on the three
diverse scenarios from the simulated datasets of section \ref{subsec:AP-sim}.
Using a 10-fold cross validation, the average classification performance
is summarized in Fig. \ref{fig:Sim_data_classification}. Observe
that in dataset (i), where features of the PPs from one cluster are
well separated from those of the other clusters, all distances perform
well. In dataset (ii) and (iii), where features of the PPs from one
cluster overlap with those of the other clusters, the OSPA distance
outperforms the Hausdorff and Wasserstein since it can appropriately
capture the cardinality dissimilarities in the data (Fig. \ref{fig:Sim_data_classification}).

\subsubsection{Classification of Texture data\label{subsec:k-NN-Texture}}

This experiment examines the classification of the extracted PP data
in section \ref{subsec:AP-Texture}, consisting of three classes from
the Texture images dataset. Using a 4-fold cross validation, the average
performance is summarized in Fig. \ref{fig:Texture_kNN_classification}.
Observe that in this dataset, the OSPA distance also performs best,
since it can give a good balance between feature and cardinality dissimilarities.

\subsubsection{Classification of StudentLife data\label{subsec:k-NN-StudentLife}}

\begin{figure}[tbh]
\begin{centering}
\vspace{-4mm}
\par\end{centering}
\begin{centering}
\subfloat[\label{fig:Texture_kNN_classification}]{\centering{}\includegraphics[width=36mm]{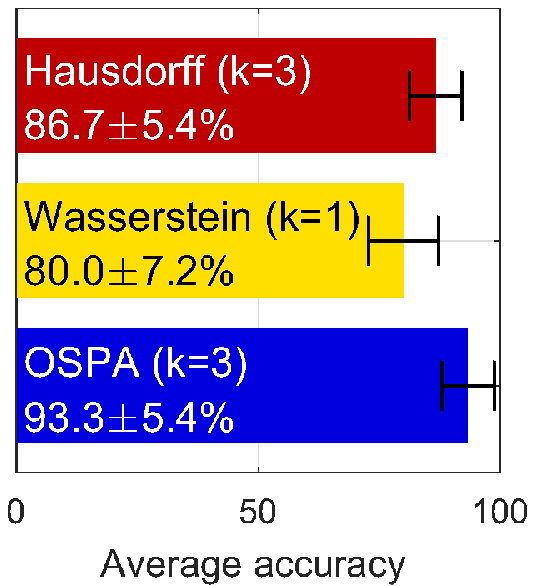}\vspace{-0mm}
}\subfloat[\label{fig:StudentLife_kNN_classification}]{\begin{centering}
\includegraphics[width=36mm]{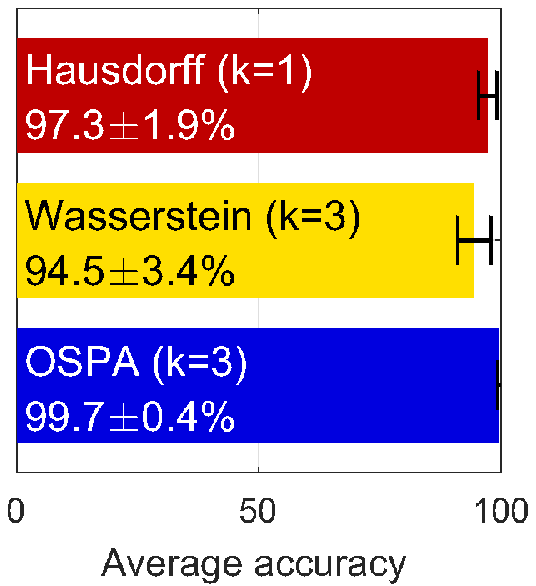}\vspace{-0mm}
\par\end{centering}
}
\par\end{centering}
\caption{Classification performance on (a) Texture data, (b) StudentLife data,
for various distances. }

\vspace{-1mm}
\end{figure}

This experiment examines the classification of the StudentLife WiFi
dataset of section \ref{subsec:AP_studentlife}. Using a 10-fold cross
validation, the average performance is summarized in Fig. \ref{fig:StudentLife_kNN_classification}.
For this dataset, all the Hausdorff, Wasserstein and OSPA achieve
good performance, since the features (i.e., WiFi IDs) from the PPs
of each cluster are well-separated from those of the other clusters. 

\section{Novelty Detection for Point Patterns\label{sec:Novelty-detection}}

Novelty detection is the task of identifying new or strange data that
are significantly different from `normal' training data \cite{markou2003novelty_p1,pimentel2014review_novel_detect}.
Note that novelty detection is not a special case of classification
because anomalous or novel training data is not available \cite{hodge2004survey}.
There are typically two phases in novelty detection: training and
detection. Since its training phase requires \emph{only normal data},
novelty detection is considered as semi-supervised learning \cite{chandola2009anomaly,hodge2004survey}.
Novelty detection is a fundamental problem in data analysis with a
plethora of application areas ranging from intrusion detection \cite{garfinkel2003Intrusion_Detection},
fraud detection \cite{bolton2002fraud_detection}, structural health
monitoring \cite{farrar2007health_monitoring}, to tumor detection
from MRI images \cite{chandola2009anomaly}. However, novelty detection
for point pattern data has not been studied.

This section introduces a solution to the novelty detection problem
for PP data by incorporating set distances into nearest neighbour
algorithm. Like classification, novelty detection can be approached
with or without knowledge of the underlying data model. The most common
non-parametric novelty detection technique is nearest neighbour \cite{pimentel2014review_novel_detect},
which is based on the assumption that normal observations are closer
to the training data than novelties \cite{hautamaki2004outlier_knn_graph}.
This approach requires a suitable notion of distance between observations
\cite{pimentel2014review_novel_detect}.

\subsection{Novelty detection with set distances}

If the distance (e.g., Hausdorff, Wasserstein or OSPA) between the
candidate PP and its nearest normal neighbour (NNN)\footnote{This can be interpreted as the Hausdorff distance between the candidate
and the normal data class.} is \emph{greater than a given threshold}, then the candidate is deemed
a novelty, otherwise it is normal. A suitable threshold can be chosen
experimentally \cite{markou2003novelty_p1}. One suitable threshold
is the 95\textsuperscript{th}-percentile of the inter-class distances
(between normal training observations and their NNNs). However, no
single threshold is guaranteed to work well for all cases. 

Similar to classification with OSPA (section \ref{subsec:k-NN-classification-with-set}),
training data can be used to determine a suitable balance between
\textit{\emph{feature dissimilarity and cardinality dissimilarit}}y.
However, there is no inter-class dissimilarity, and hence minimizing
the intra-class dissimilarity for normal data yields the trivial solution
$c=0$. To determine a suitable balance, consider the cardinality
dissimilarity $d_{\mathtt{card}}^{(p)}(X,Y)=\frac{1}{n}\left(n-m\right)$
and feature dissimilarity $d_{\mathtt{feat}}^{(p)}(X,Y)=\frac{1}{n}\min_{\pi\in\Pi_{n}}\sum_{i=1}^{m}\underline{d}\left(x_{i},y_{\pi(i)}\right)^{p}$
between all pairs of observations $X,Y$ in the normal training set
(assuming the cardinality $m$ of $Y$ is not greater than the cardinality
$n$ of $X$, otherwise we compute $d_{\mathtt{card}}^{(p)}(Y,X)$
and $d_{\mathtt{feat}}^{(p)}(Y,X)$). Note that for $d_{\mathtt{feat}}^{(p)}$
we use the base distance $\underline{d}$ to capture the absolute
feature dissimilarity rather than the capped feature dissimilarity
from base distance $\underline{d}^{(c)}$. To decide whether a test
PP $T$ is novel, we need to determine its cardinality dissmilarity
and feature dissimilarity relative to the normal data. The relative
cardinality dissmilarity and feature dissimilarity of $T$ (with respect
to the normal data) can be defined as $d_{\mathtt{card}}^{(p)}(T,T^{*}))/m_{\mathtt{card}}^{(p)}$
and $d_{\mathtt{feat}}^{(p)}(T,T^{*})/m_{\mathtt{feat}}^{(p)}$, where
$T^{*}$ is $T$'s NNN, $m_{\mathtt{card}}^{(p)}$ and $m_{\mathtt{feat}}^{(p)}$
are large values (e.g., maximum or 95\textsuperscript{th}-percentile)
of $d_{\mathtt{card}}^{(p)}(Y,X)$ and $d_{\mathtt{feat}}^{(p)}(Y,X)$)
in the normal data set, respectively. Observe that summing the relative
dissmilarities and scaling by $m_{\mathtt{feat}}^{(p)}$ gives the
uncapped OSPA ``distance''
\[
(d_{\mathtt{O}}^{(p)}(T,X(T)))^{p}=\frac{m_{\mathtt{feat}}^{(p)}}{m_{\mathtt{card}}^{(p)}}d_{\mathtt{card}}^{(p)}((T,T^{*})+d_{\mathtt{feat}}^{(p)}(T,T^{*})
\]
Hence, a suitable cut-off parameter is $c=\left(m_{\mathtt{feat}}^{(p)}/m_{\mathtt{card}}^{(p)}\right)^{1/p}$
.

\subsection{Experiments\label{subsec:k-NN-novel-exp}}

In this section, we examine the novelty detection performance of the
Hausdorff, Wasserstein and OSPA  distances on both simulated and real
data. 

\subsubsection{Novelty detection with simulated data\label{subsec:k-NN-novel-sim}}

In this experiment, we consider cluster 2 from the simulated data
set in subsection \ref{subsec:AP-sim} as normal data, and clusters
1 and 3 as novel data. This allows us to study three diverse scenarios:
dataset (i), see Fig. \ref{fig:AP_sim_sepa_feat}, is an example of\emph{
feature novelty,} where novel observations are similar in cardinality
with normal training data, but dissimilar in feature; dataset (ii),
shown in Fig. \ref{fig:AP_sim_sepa_card}, is an example of \emph{cardinality
novelty}, where novel observations are similar in feature with normal
training data, but dissimilar in cardinality; dataset (iii), shown
in Fig. \ref{fig:AP_sim_mix}, is a mix of feature and cardinality
novelty.

\begin{figure}[tbh]
\begin{centering}
\vspace{-0mm}
\par\end{centering}
\begin{centering}
\hspace{-3mm}\includegraphics[width=90mm]{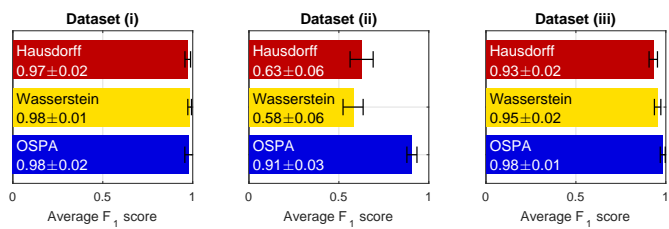}\vspace{-0mm}
\par\end{centering}
\caption{\label{fig:Sim_data_novelty}Novelty detection performance on simulated
data for various distances. }

\vspace{-1mm}
\end{figure}

\begin{figure}[tbh]
\begin{centering}
\vspace{-2mm}
\subfloat[\label{fig:Sim_anomaly_boxplot_data_1}Dataset (i)]{\begin{centering}
\hspace{-3mm}\includegraphics[width=90mm]{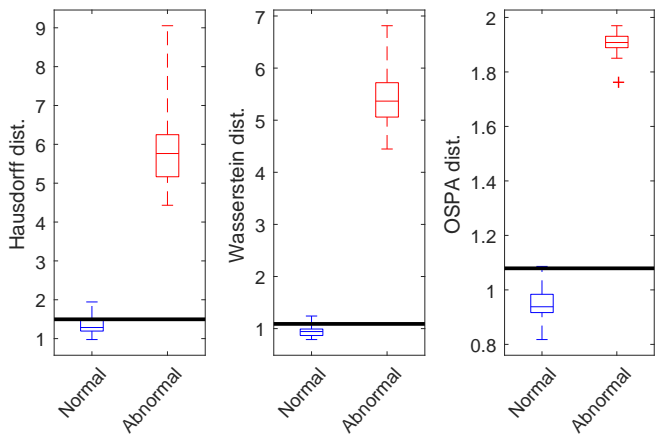}
\par\end{centering}
\vspace{-5mm}
}
\par\end{centering}
\begin{centering}
\vspace{-0mm}
\subfloat[\label{fig:Sim_anomaly_boxplot_data_2}Dataset (ii)]{\begin{centering}
\hspace{-3mm}\includegraphics[width=90mm]{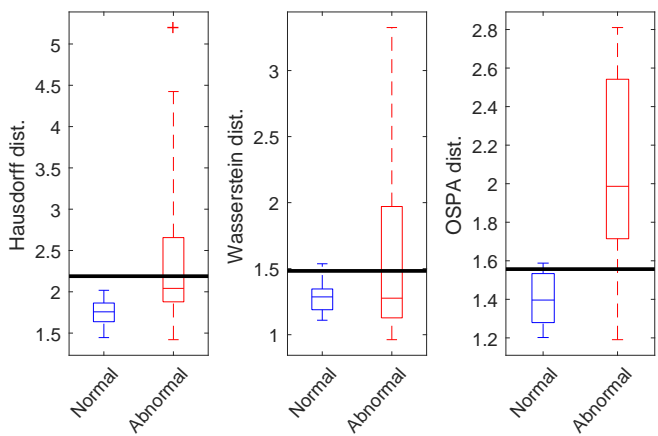}
\par\end{centering}
\vspace{-5mm}
}
\par\end{centering}
\begin{centering}
\vspace{-0mm}
\subfloat[\label{fig:Sim_anomaly_boxplot_data_3}Dataset (iii)]{\begin{centering}
\hspace{-3mm}\includegraphics[width=90mm]{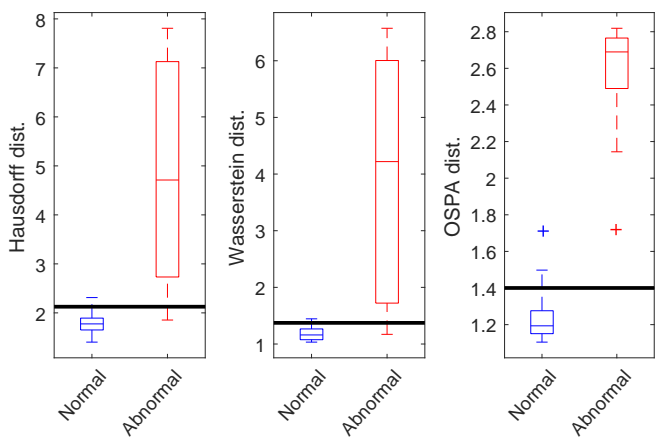}
\par\end{centering}
\vspace{-5mm}
}\vspace{-0mm}
\par\end{centering}
\caption{\label{fig:Sim_novelty_boxplots}Boxplots of distances between test
data and their NNNs for one data fold of the simulated dataset. Thick
lines across the boxes indicate the chosen thresholds.}
\vspace{-4mm}
\end{figure}

Using a 10-fold cross validation, the average performance summarized
in Fig. \ref{fig:Sim_data_novelty}. Fig. \ref{fig:Sim_novelty_boxplots}
shows boxplots of the distances between the test PPs and theirs NNNs. Observe that in datasets (i) and
(iii), where novelties are dissimilar with normal data in feature
(see Figs. \ref{fig:AP_sim_sepa_feat} and \ref{fig:AP_sim_mix}),
all distances perform well. In dataset (ii), where novelties are dissimilar
with normal data in cardinality, but similar in feature (see Fig.
\ref{fig:AP_sim_sepa_card}), the OSPA distance outperforms the Hausdorff
and Wasserstein since it can appropriately penalize the cardinality
dissimilarity between normal and novel data (see Fig. \ref{fig:Sim_anomaly_boxplot_data_2}).

\subsubsection{Novelty detection with Texture data\label{subsec:k-NN-novel-Texture}}

Using the Texture dataset from subsection \ref{subsec:AP-Texture},
we consider normal data are taken from class ``T14 brick1'' and
novel data are taken from class ``T20 upholstery''. We use 4-fold
cross validation. In each fold, the training data consist of 75\%
of images from normal class (30 images), the testing set includes
the remaining images from normal class (10 images) and 25\% of images
from novel class (10 images). 

\begin{figure}[tbh]
\begin{centering}
\vspace{-3mm}
\hspace{-3mm}\includegraphics[width=37mm]{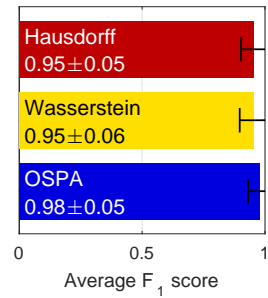}
\par\end{centering}
\vspace{-0mm}
\caption{\label{fig:Texture_kNN_novelty}Novelty detection performance on the
Texture data for various distances. }

\vspace{-2mm}
\end{figure}

Observe that the performance of set distances (Hausdorff, Wasserstein,
and OSPA) on this dataset is similar to that of set distances on the
simulated dataset in section \ref{subsec:k-NN-novel-sim}. Since normal
and novel data are dissimilar in feature (see the feature plot in
Fig. \ref{fig:Texture_data_results}), all distances perform well
(Fig. \ref{fig:Texture_kNN_novelty}). 

\begin{figure}[tbh]
\begin{centering}
\vspace{-0mm}
\hspace{-3mm}\includegraphics[width=90mm]{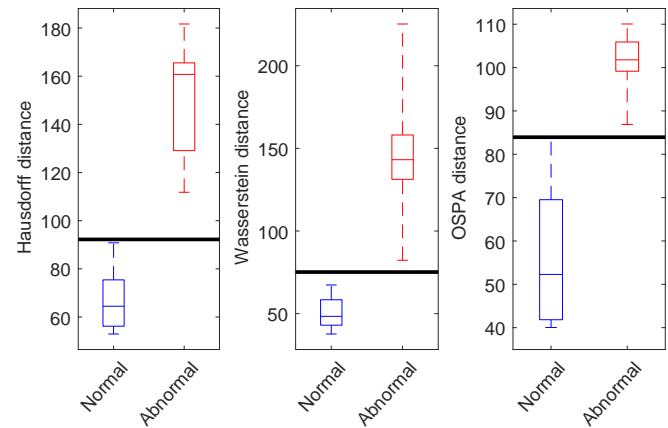}
\par\end{centering}
\vspace{-0mm}
\caption{\label{fig:Texture_novelty_boxplot}Boxplots of various NNN distances
for one data fold of the Texture dataset. Thick lines across the boxes
indicate the chosen thresholds.}
\vspace{-2mm}
\end{figure}

\subsubsection{Novelty detection with StudentLife WiFi data\label{subsec:k-NN-novel-StudentLife}}

\begin{figure}[tbh]
\begin{centering}
\vspace{-1mm}
\includegraphics[width=37mm]{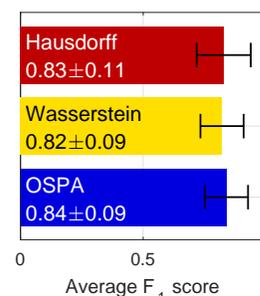}\vspace{-2mm}
\par\end{centering}
\caption{\label{fig:StudentLife-anomaly_result}Novelty detection performance
on the StudentLife data for various set distances. }
\vspace{-3mm}
\end{figure}

Using the StudentLife WiFi dataset described in subsection \ref{subsec:AP_studentlife},
we consider observations from locations 1 and 2 as normal data and
observations from locations 3 and 4 as novelties. Using a 10-fold
cross validation, the average performance is summarized in Fig. \ref{fig:StudentLife-anomaly_result}.
Observe that all three distances  (OSPA, Hausdorff and Wasserstein
distance) perform similar for this dataset with average F\textsubscript{1}
score about 0.83.

\begin{figure}[tbh]
\begin{centering}
\vspace{-0mm}
\hspace{-3mm}\includegraphics[width=90mm]{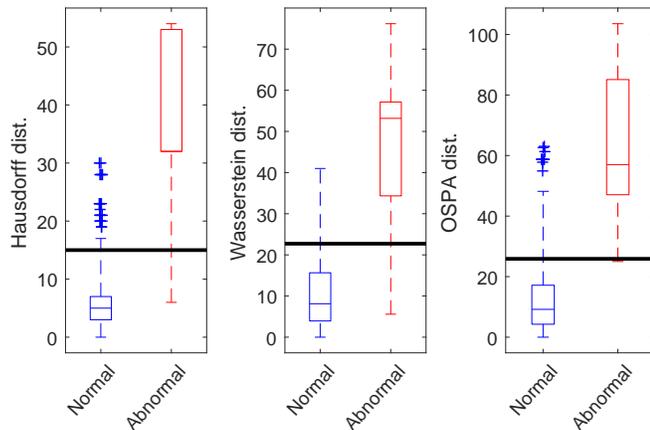}
\par\end{centering}
\vspace{-1mm}
\caption{\label{fig:StudentLife-Boxplot}Boxplots of distances between test
data and their NNNs for one data fold of the StudentLife dataset.
Thick lines across the boxes indicate the chosen thresholds. }
\vspace{-2mm}
\end{figure}

\section{Conclusions\label{sec:Conclusions}}

In this paper, algorithms for clustering, classification, and novelty
detection with point pattern data using the OSPA distance have been
presented. In clustering, AP is combined with the OSPA (or others
such as Wasserstein and Hausdorff) distance as dissimilarity measure.
In classification, the OSPA distance is incorporated into the $k$-nearest
neighbour ($k$-NN) algorithm. In MI novelty detection, a solution
developed using the set distances between the candidate PP and its
nearest normal neighbour in the training set. 

Numerical experiments on simulated and real data demonstrated that
the OSPA distance offers more flexibility in design choices as well
as the ability to better capture dissimilarities between sets compared
to the other distances. We reiterate that while the OSPA distance
does offer some merits over the other distances, there is no single
distance that works for all applications. In practice, to determine
which distance (and parameters) are better suited to which application,
it is important to assess whether the distance agrees with the notion
of similarity/dissimilarity specific to that application.

\vspace{-1mm}

\bibliographystyle{IEEEtran}
\bibliography{Refs/ref1,Refs/ref_BaNgu,Refs/ref_BaTuong,Refs/ref_Dinh,Refs/ref_Thuong}

\begin{thebibliography}{10}
\providecommand{\url}[1]{#1}
\csname url@samestyle\endcsname
\providecommand{\newblock}{\relax}
\providecommand{\bibinfo}[2]{#2}
\providecommand{\BIBentrySTDinterwordspacing}{\spaceskip=0pt\relax}
\providecommand{\BIBentryALTinterwordstretchfactor}{4}
\providecommand{\BIBentryALTinterwordspacing}{\spaceskip=\fontdimen2\font plus
\BIBentryALTinterwordstretchfactor\fontdimen3\font minus
  \fontdimen4\font\relax}
\providecommand{\BIBforeignlanguage}[2]{{%
\expandafter\ifx\csname l@#1\endcsname\relax
\typeout{** WARNING: IEEEtran.bst: No hyphenation pattern has been}%
\typeout{** loaded for the language `#1'. Using the pattern for}%
\typeout{** the default language instead.}%
\else
\language=\csname l@#1\endcsname
\fi
#2}}
\providecommand{\BIBdecl}{\relax}
\BIBdecl

\bibitem{dietterich1997solving_multiple_instance}
T.~G. Dietterich, R.~H. Lathrop, and T.~Lozano-P{\'e}rez, ``Solving the
  multiple instance problem with axis-parallel rectangles,'' \emph{Artificial
  intelligence}, vol.~89, no.~1, pp. 31--71, 1997.

\bibitem{minhas2012multiple}
F.~Minhas and A.~Ben-Hur, ``Multiple instance learning of calmodulin binding
  sites,'' \emph{Bioinformatics (Oxford, England)}, vol.~28, no.~18, pp.
  i416--i422, 2012.

\bibitem{amores2013multiple_intance_review}
J.~Amores, ``Multiple instance classification: Review, taxonomy and comparative
  study,'' \emph{Artificial Intelligence}, vol. 201, pp. 81--105, 2013.

\bibitem{foulds2010multi_instance_review}
J.~Foulds and E.~Frank, ``A review of multi-instance learning assumptions,''
  \emph{The Knowledge Engineering Review}, vol.~25, no.~01, pp. 1--25, 2010.

\bibitem{moller2003point_processes}
J.~Moller and R.~P. Waagepetersen, \emph{Statistical inference and simulation
  for spatial point processes}.\hskip 1em plus 0.5em minus 0.4em\relax CRC
  Press, 2003.

\bibitem{joachims1996probabilistic}
T.~Joachims, ``A probabilistic analysis of the rocchio algorithm with tfidf for
  text categorization.'' DTIC Document, Tech. Rep., 1996.

\bibitem{mccallum1998comparison_NBtextClassifi}
A.~McCallum and K.~Nigam, ``{A comparison of event models for naive Bayes text
  classification},'' in \emph{AAAI-98 Workshop learning for text
  categorization}, vol. 752.\hskip 1em plus 0.5em minus 0.4em\relax Citeseer,
  1998, pp. 41--48.

\bibitem{csurka2004visual}
G.~Csurka, C.~Dance, L.~Fan, J.~Willamowski, and C.~Bray, ``Visual
  categorization with bags of keypoints,'' in \emph{Workshop statistical
  learning in computer vision, ECCV}, 2004.

\bibitem{fei2005bayesian}
L.~Fei-Fei and P.~Perona, ``{A Bayesian hierarchical model for learning natural
  scene categories},'' in \emph{IEEE Comput. Soc. Conf. Comput. Vision and
  Pattern Recognition (CVPR), 2005}, vol.~2.\hskip 1em plus 0.5em minus
  0.4em\relax IEEE, 2005, pp. 524--531.

\bibitem{rusu2011point_cloud_lib}
R.~B. Rusu and S.~Cousins, ``3d is here: Point cloud library (pcl),'' in
  \emph{IEEE Int. Conf. Robotics and Automation (ICRA), 2011}.\hskip 1em plus
  0.5em minus 0.4em\relax IEEE, 2011, pp. 1--4.

\bibitem{sitek2006tomographic_point_cloud}
A.~Sitek, R.~H. Huesman, and G.~T. Gullberg, ``Tomographic reconstruction using
  an adaptive tetrahedral mesh defined by a point cloud,'' \emph{IEEE Trans.
  Medical Imaging}, vol.~25, no.~9, pp. 1172--1179, 2006.

\bibitem{woo2002segmentation_point_cloud}
H.~Woo, E.~Kang, S.~Wang, and K.~H. Lee, ``A new segmentation method for point
  cloud data,'' \emph{Int. Journal Machine Tools and Manufacture}, vol.~42,
  no.~2, pp. 167--178, 2002.

\bibitem{guha1999rock}
S.~Guha, R.~Rastogi, and K.~Shim, ``Rock: A robust clustering algorithm for
  categorical attributes,'' in \emph{Proc. 15th Int. Conf. Data Eng.,
  1999}.\hskip 1em plus 0.5em minus 0.4em\relax IEEE, 1999, pp. 512--521.

\bibitem{yang2002clope}
Y.~Yang, X.~Guan, and J.~You, ``Clope: a fast and effective clustering
  algorithm for transactional data,'' in \emph{Proc. 8th ACM SIGKDD Int. Conf.
  Knowledge Discovery and Data Mining}.\hskip 1em plus 0.5em minus 0.4em\relax
  ACM, 2002, pp. 682--687.

\bibitem{yun2001clustering_basket_data}
C.-H. Yun, K.-T. Chuang, and M.-S. Chen, ``An efficient clustering algorithm
  for market basket data based on small large ratios,'' in \emph{Comput. Softw.
  and Appl. Conf., 2001. COMPSAC 2001. 25th Annual Int.}\hskip 1em plus 0.5em
  minus 0.4em\relax IEEE, 2001, pp. 505--510.

\bibitem{cadez2000EMclustering_VariableLengthData}
I.~V. Cadez, S.~Gaffney, and P.~Smyth, ``A general probabilistic framework for
  clustering individuals and objects,'' in \emph{Proc. 6th ACM SIGKDD Int.
  Conf. knowledge discovery and data mining}.\hskip 1em plus 0.5em minus
  0.4em\relax ACM, 2000, pp. 140--149.

\bibitem{hodge2004survey}
V.~J. Hodge and J.~Austin, ``A survey of outlier detection methodologies,''
  \emph{Artificial Intelligence Review}, vol.~22, no.~2, pp. 85--126, 2004.

\bibitem{zhang2009MIClustering}
M.-L. Zhang and Z.-H. Zhou, ``Multi-instance clustering with applications to
  multi-instance prediction,'' \emph{Appl. Intell.}, vol.~31, no.~1, pp.
  47--68, 2009.

\bibitem{zhang2009m3icClustering}
D.~Zhang, F.~Wang, L.~Si, and T.~Li, ``M3ic: Maximum margin multiple instance
  clustering,'' in \emph{IJCAI}, vol.~9, 2009, pp. 1339--1344.

\bibitem{huttenlocher1993tracking_Hausdorff}
D.~P. Huttenlocher, J.~J. Noh, and W.~J. Rucklidge, ``Tracking non-rigid
  objects in complex scenes,'' in \emph{Proc. 4th Int. Conf. Comput. Vision,
  1993}.\hskip 1em plus 0.5em minus 0.4em\relax IEEE, 1993, pp. 93--101.

\bibitem{gavrila1999Chamfer_distance}
D.~M. Gavrila and V.~Philomin, ``Real-time object detection for "smart"
  vehicles,'' in \emph{Proc. 7th Int. Conf. Comput. Vision, 1999},
  vol.~1.\hskip 1em plus 0.5em minus 0.4em\relax IEEE, 1999, pp. 87--93.

\bibitem{zhang2007EMD_kernel}
J.~Zhang, M.~Marsza{\l}ek, S.~Lazebnik, and C.~Schmid, ``Local features and
  kernels for classification of texture and object categories: A comprehensive
  study,'' \emph{Int. J. Comput. Vision}, vol.~73, no.~2, pp. 213--238, 2007.

\bibitem{rubner1998EarthMoversDistance}
Y.~Rubner, C.~Tomasi, and L.~J. Guibas, ``A metric for distributions with
  applications to image databases,'' in \emph{6th Int. Conf. Comput. Vision,
  1998}.\hskip 1em plus 0.5em minus 0.4em\relax IEEE, 1998, pp. 59--66.

\bibitem{schuhmacher2008_OSPA}
D.~Schuhmacher, B.-T. Vo, and B.-N. Vo, ``A consistent metric for performance
  evaluation of multi-object filters,'' \emph{IEEE Trans. Signal Process.},
  vol.~56, no.~8, pp. 3447--3457, 2008.

\bibitem{frey2007_APclustering}
B.~J. Frey and D.~Dueck, ``Clustering by passing messages between data
  points,'' \emph{Sci.}, vol. 315, no. 5814, pp. 972--976, 2007.

\bibitem{tran2016clustering_PP}
Q.~N. Tran, B.-N. Vo, D.~Phung, and B.-T. Vo, ``Clustering for point pattern
  data,'' in \emph{23rd Intl. Conf. Pattern Recognition (ICPR)}, Dec. 2016.

\bibitem{cover1967nearest_neighbor_classifier}
T.~M. Cover and P.~E. Hart, ``Nearest neighbor pattern classification,''
  \emph{IEEE Trans. Inf. Theory}, vol.~13, no.~1, pp. 21--27, 1967.

\bibitem{keller1985fuzzy_kNN}
J.~M. Keller, M.~R. Gray, and J.~A. Givens, ``A fuzzy k-nearest neighbor
  algorithm,'' \emph{IEEE Trans. Systems, Man and Cybernetics}, no.~4, pp.
  580--585, 1985.

\bibitem{hoffman2004multitarget_distance}
J.~R. Hoffman and R.~P. Mahler, ``Multitarget miss distance via optimal
  assignment,'' \emph{IEEE Trans. Systems, Man and Cybernetics, Part A: Systems
  and Humans}, vol.~34, no.~3, pp. 327--336, 2004.

\bibitem{jain2010clustering50yearsKmeans}
A.~K. Jain, ``Data clustering: 50 years beyond k-means,'' \emph{Pattern
  recognition letters}, vol.~31, no.~8, pp. 651--666, 2010.

\bibitem{pimentel2014review_novel_detect}
M.~A. Pimentel, D.~A. Clifton, L.~Clifton, and L.~Tarassenko, ``A review of
  novelty detection,'' \emph{Signal Process.}, vol.~99, pp. 215--249, 2014.

\bibitem{huttenlocher1993comparing}
D.~P. Huttenlocher, G.~A. Klanderman, and W.~J. Rucklidge, ``Comparing images
  using the hausdorff distance,'' \emph{IEEE Trans. Pattern Anal. Mach.
  Intell.}, vol.~15, no.~9, pp. 850--863, 1993.

\bibitem{rucklidge1995locating_obj}
W.~J. Rucklidge, ``Locating objects using the hausdorff distance,'' in
  \emph{Proc. 5th Int. Conf. Comput. Vision, 1995}.\hskip 1em plus 0.5em minus
  0.4em\relax IEEE, 1995, pp. 457--464.

\bibitem{cignoni1998metro}
P.~Cignoni, C.~Rocchini, and R.~Scopigno, ``Metro: measuring error on
  simplified surfaces,'' in \emph{Comput. Graphics Forum}, vol.~17,
  no.~2.\hskip 1em plus 0.5em minus 0.4em\relax Wiley Online Library, 1998, pp.
  167--174.

\bibitem{baddeley1992errors}
A.~Baddeley, ``Errors in binary images and an lp version of the hausdorff
  metric,'' \emph{Nieuw Archief voor Wiskunde}, vol.~10, no.~4, pp. 157--183,
  1992.

\bibitem{Jain1999data_clustering}
A.~K. Jain, M.~N. Murty, and P.~J. Flynn, ``Data clustering: a review,''
  \emph{ACM Comput. surveys (CSUR)}, vol.~31, no.~3, pp. 264--323, 1999.

\bibitem{russell2003artificial}
S.~Russell and P.~Norvig, \emph{Artificial Intelligence: A modern
  approach}.\hskip 1em plus 0.5em minus 0.4em\relax Prentice Hall, 2003.

\bibitem{murphy2012machine}
K.~P. Murphy, \emph{Machine learning: a probabilistic perspective}.\hskip 1em
  plus 0.5em minus 0.4em\relax MIT press, 2012.

\bibitem{tryon1939cluster}
R.~C. Tryon, \emph{Cluster analysis: correlation profile and orthometric
  (factor) analysis for the isolation of unities in mind and
  personality}.\hskip 1em plus 0.5em minus 0.4em\relax Edwards brothers, 1939.

\bibitem{xu2005survey_clustering}
R.~Xu and D.~Wunsch, ``Survey of clustering algorithms,'' \emph{IEEE Trans.
  Neural Networks}, vol.~16, no.~3, pp. 645--678, 2005.

\bibitem{baum2015ospa_clustering}
M.~Baum, B.~Balasingam, P.~Willett, and U.~D. Hanebeck, ``Ospa barycenters for
  clustering set-valued data,'' in \emph{18th Int. Conf. Inf. Fusion
  (Fusion)}.\hskip 1em plus 0.5em minus 0.4em\relax IEEE, 2015, pp. 1375--1381.

\bibitem{dueck2007non_metric_AP}
D.~Dueck and B.~J. Frey, ``Non-metric affinity propagation for unsupervised
  image categorization,'' in \emph{11th Int. Conf. Comput. Vision
  (ICCV)}.\hskip 1em plus 0.5em minus 0.4em\relax IEEE, 2007, pp. 1--8.

\bibitem{givoni2009binary_AP}
I.~E. Givoni and B.~J. Frey, ``A binary variable model for affinity
  propagation,'' \emph{Neural computation}, vol.~21, no.~6, pp. 1589--1600,
  2009.

\bibitem{manning2008info_retrieval}
C.~D. Manning, P.~Raghavan, and H.~Sch{\"u}tze, \emph{Introduction to
  information retrieval}.\hskip 1em plus 0.5em minus 0.4em\relax Cambridge
  univ. press Cambridge, 2008, vol.~1.

\bibitem{vo2016model-based_PP}
B.-N. Vo, Q.~N. Tran, D.~Phung, and B.-T. Vo, ``Model-based classification and
  novelty detection for point pattern data,'' in \emph{23rd Intl. Conf. Pattern
  Recognition (ICPR)}, Dec. 2016.

\bibitem{textureDataset}
S.~Lazebnik, C.~Schmid, and J.~Ponce, ``A sparse texture representation using
  local affine regions,'' \emph{IEEE Trans. Pattern Anal. Mach. Intell.},
  vol.~27, no.~8, pp. 1265--1278, 2005.

\bibitem{vlfeatLib}
A.~Vedaldi and B.~Fulkerson, ``Vlfeat: An open and portable library of comput.
  vision algorithms,'' http://www.vlfeat.org/, 2008.

\bibitem{studentlife_dataset}
R.~Wang, F.~Chen, Z.~Chen, T.~Li, G.~Harari, S.~Tignor, X.~Zhou, D.~Ben-Zeev,
  and A.~T. Campbell, ``{StudentLife: assessing mental health, academic
  performance and behavioral trends of college students using smartphones},''
  in \emph{Proc. 2014 ACM Int. Joint Conf. Pervasive and Ubiquitous
  Comput.}\hskip 1em plus 0.5em minus 0.4em\relax ACM, 2014, pp. 3--14.

\bibitem{bishop2006pattern}
C.~M. Bishop, \emph{Pattern recognition and machine learning}.\hskip 1em plus
  0.5em minus 0.4em\relax Springer, 2006.

\bibitem{witten2005data_mining}
I.~H. Witten and E.~Frank, \emph{Data Mining: Practical machine learning tools
  and techniques}.\hskip 1em plus 0.5em minus 0.4em\relax Morgan Kaufmann,
  2005.

\bibitem{boser1992training_svm}
B.~E. Boser, I.~M. Guyon, and V.~N. Vapnik, ``A training algorithm for optimal
  margin classifiers,'' in \emph{Proc. 5th Annual Workshop Computational
  Learning Theory}.\hskip 1em plus 0.5em minus 0.4em\relax ACM, 1992, pp.
  144--152.

\bibitem{cortes1995SVM}
C.~Cortes and V.~Vapnik, ``Support-vector networks,'' \emph{Machine learning},
  vol.~20, no.~3, pp. 273--297, 1995.

\bibitem{duda2001pattern}
R.~O. Duda, P.~E. Hart, and D.~G. Stork, \emph{Pattern classification (2nd
  edition)}.\hskip 1em plus 0.5em minus 0.4em\relax John Wiley \& Sons, 2001.

\bibitem{markou2003novelty_p1}
M.~Markou and S.~Singh, ``Novelty detection: a review -- part 1: statistical
  approaches,'' \emph{Signal Process.}, vol.~83, no.~12, pp. 2481--2497, 2003.

\bibitem{chandola2009anomaly}
V.~Chandola, A.~Banerjee, and V.~Kumar, ``Anomaly detection: A survey,''
  \emph{ACM Comput. Surveys (CSUR)}, vol.~41, no.~3, p.~15, 2009.

\bibitem{garfinkel2003Intrusion_Detection}
T.~Garfinkel, M.~Rosenblum \emph{et~al.}, ``A virtual machine introspection
  based architecture for intrusion detection,'' in \emph{NDSS}, vol.~3, 2003,
  pp. 191--206.

\bibitem{bolton2002fraud_detection}
R.~J. Bolton and D.~J. Hand, ``Statistical fraud detection: A review,''
  \emph{Statistical Sci.}, pp. 235--249, 2002.

\bibitem{farrar2007health_monitoring}
C.~R. Farrar and K.~Worden, ``An introduction to structural health
  monitoring,'' \emph{Philosophical Trans. Royal Soc. London A: Mathematical,
  Physical and Engineering Sci.}, vol. 365, no. 1851, pp. 303--315, 2007.

\bibitem{hautamaki2004outlier_knn_graph}
V.~Hautam{\"a}ki, I.~K{\"a}rkk{\"a}inen, and P.~Fr{\"a}nti, ``Outlier detection
  using k-nearest neighbour graph,'' in \emph{ICPR}, 2004, pp. 430--433.

\end{thebibliography}

\end{document}